\documentclass[10pt,journal,compsoc]{IEEEtran}
\usepackage{amsmath,amsfonts}
\usepackage{algorithmic}
\usepackage{array}
\usepackage[caption=false,font=normalsize,labelfont=sf,textfont=sf]{subfig}
\usepackage{textcomp}
\usepackage{stfloats}
\usepackage{url}
\usepackage{verbatim}
\usepackage{graphicx}
\usepackage{cite}
\usepackage{multirow}
\usepackage{wrapfig}
\usepackage{booktabs}
\usepackage{colortbl}
\usepackage{hyperref}
\usepackage{threeparttable}
\hypersetup{
	colorlinks=true,
	citecolor=blue,
	filecolor=red,
	urlcolor=magenta,
}
\newcommand*{\pcr}{\fontfamily{pcr}\selectfont}
\definecolor{ggray}{rgb}{0.90, 0.90, 0.98}
\begin{document}

% \title{A Sample Article Using IEEEtran.cls\\ for IEEE Journals and Transactions}
\title{A2Mamba: Attention-augmented State Space Models for Visual Recognition}

\author{Meng Lou, Yunxiang Fu, and Yizhou Yu,~\IEEEmembership{Fellow,~IEEE}
        % <-this % stops a space
\thanks{Meng Lou, Yunxiang Fu, and Yizhou Yu are with the School of Computing and Data Science, The University of Hong Kong, Hong Kong SAR, China. (E-mail: {loumeng@connect.hku.hk}; {yunxiang@connect.hku.hk}; {yizhouy@acm.org})}% <-this % stops a space
% \thanks{Corresponding author: Yizhou Yu}
}

\markboth{IEEE Transactions on Pattern Analysis and Machine Intelligence}%
{Shell \MakeLowercase{\textit{et al.}}: A Sample Article Using IEEEtran.cls for IEEE Journals}


% \maketitle
% \IEEEdisplaynontitleabstractindextext

\IEEEtitleabstractindextext{
\begin{abstract}
Transformers and Mamba, initially invented for natural language processing, have inspired backbone architectures for visual recognition. Recent studies integrated Local Attention Transformers with Mamba to capture both local details and global contexts. Despite competitive performance, these methods are limited to simple stacking of Transformer and Mamba layers without any interaction mechanism between them. Thus, deep integration between Transformer and Mamba layers remains an open problem. We address this problem by proposing A2Mamba, a powerful Transformer-Mamba hybrid network architecture, featuring a new token mixer termed Multi-scale Attention-augmented State Space Model (MASS), where multi-scale attention maps are integrated into an attention-augmented SSM (A2SSM). A key step of A2SSM performs a variant of cross-attention by spatially aggregating the SSM's hidden states using the multi-scale attention maps, which enhances spatial dependencies pertaining to a two-dimensional space while improving the dynamic modeling capabilities of SSMs. Our A2Mamba outperforms all previous ConvNet-, Transformer-, and Mamba-based architectures in visual recognition tasks. For instance, A2Mamba-L achieves an impressive 86.1\% top-1 accuracy on ImageNet-1K. In semantic segmentation, A2Mamba-B exceeds CAFormer-S36 by 2.5\% in mIoU, while exhibiting higher efficiency. In object detection and instance segmentation with Cascade Mask R-CNN, A2Mamba-S surpasses MambaVision-B by 1.2\%/0.9\% in AP$^b$/AP$^m$, while having 40\% fewer parameters. Code is publicly available at \url{https://github.com/LMMMEng/A2Mamba}.
\end{abstract}

}

\maketitle
\IEEEdisplaynontitleabstractindextext

\section{Introduction}
Vision Transformers (ViTs)~\cite{dosovitskiy2020image} have become a de-facto choice for various vision tasks due to their ability to model long-range dependencies using multi-head self-attention (MHSA) \cite{vaswani2017attention}. However, the quadratic complexity of MHSA leads to high computational costs, particularly in dense prediction tasks such as semantic segmentation and object detection, which require high-resolution inputs. To this end, subsequent efforts have proposed efficient attention mechanisms such as window attention~\cite{liu2021swin,dong2022cswin,hassani2023neighborhood,zhu2023biformer}, spatial reduction attention~\cite{wang2021pyramid,wang2021pvtv2,wu2022p2t}, and dilated attention~\cite{tu2022maxvit,hassani2022dilated,wang2023crossformer++} to reduce computational complexity. Recently, since the Mamba architecture~\cite{gu2023mamba} can model long-range dependencies with linear-time complexity, many efforts have been dedicated to developing Mamba-based architectures for visual recognition~\cite{zhu2024vision,liu2024vmamba,huang2024localmamba,yang2024plainmamba,pei2024efficientvmamba,xiao2024spatialmamba,lou2025sparx}. In contrast to spatial reduction attention and dilated attention that reduce sequence length via downsampling or shuffling, Mamba directly models long-range dependencies on the original sequence through state space models (SSMs). This architecture enables fine-grained information preservation during long-sequence processing, very promising for enabling vision models to achieve superior performance in dense prediction tasks~\cite{yu2024mambaout}.
\par
\begin{figure}[t]
    \centering
    \includegraphics[width=0.475\textwidth]{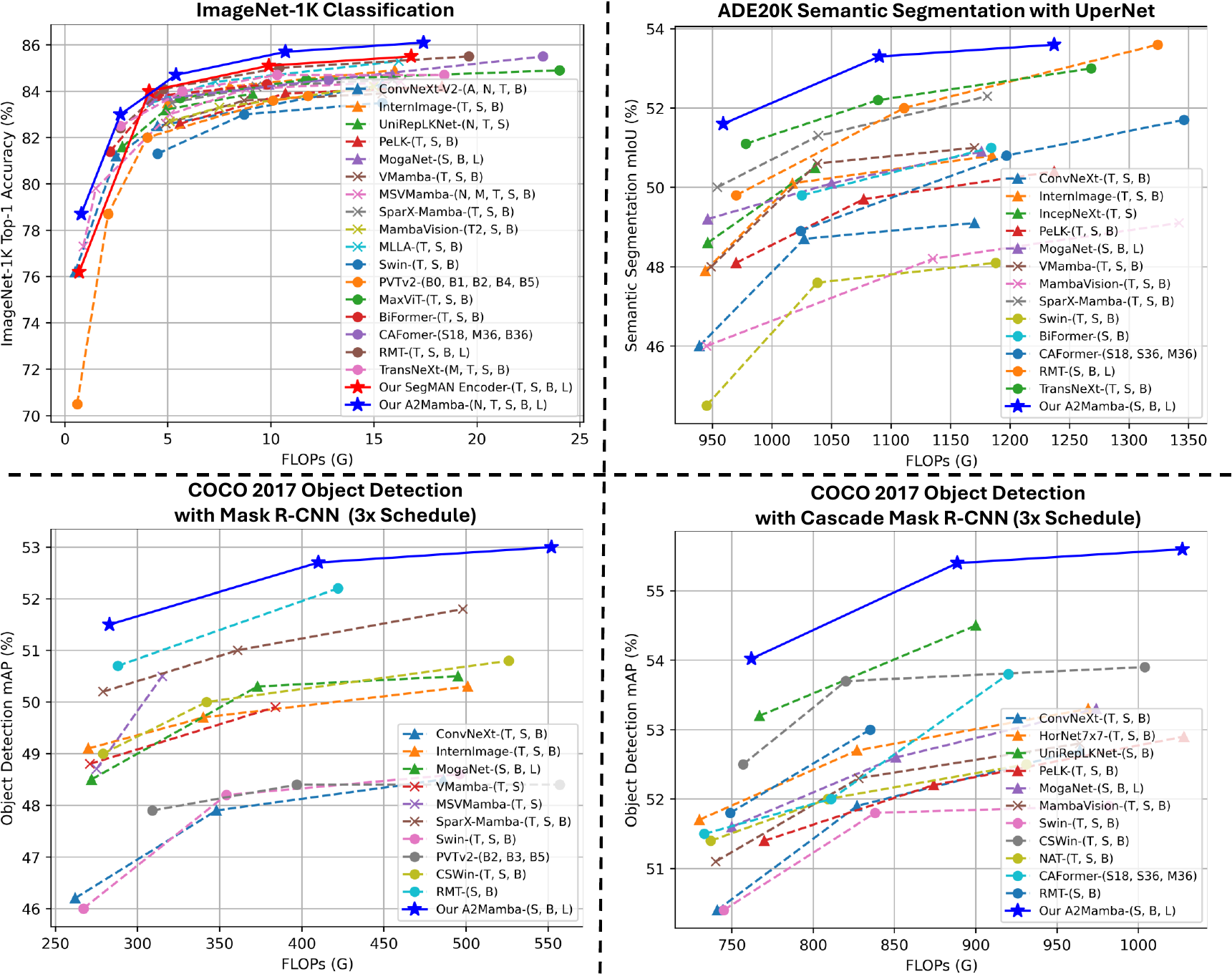}
    \caption{Performance comparisons between our A2Mamba and other representative backbone architectures on visual recognition tasks.}
    \label{fig:acc_plot}
    % \vspace{-15pt}
\end{figure}
The sequential scanning mechanism in SSMs naturally suits language modeling, where word order matters, while images exhibit complex 2D structures with non-sequential pixel dependencies. Hence, SSMs have difficulty to comprehensively understand the spatial structures of images. Although some efforts~\cite{liu2024vmamba,huang2024localmamba} have leveraged alternative scanning strategies to partially overcome this limitation, the inherent causality caused by sequential scanning still compromises latent spatial dependencies to some extent. Consequently, Transformer-Mamba hybrid architectures have emerged as a promising direction for visual recognition. For instance, MambaVision~\cite{hatamizadeh2024mambavision} constructs a vision backbone by stacking MHSA and SSM blocks in deeper stages, using MHSA to complement SSM. However, its performance still lags behind advanced ViTs~\cite{han2025mlla,metaformer2024,fan2023rmt,shi2023transnext} on diverse vision tasks despite high efficiency. Recently, a generic Transformer-Mamba hybrid architecture, termed SegMAN Encoder~\cite{fu2025segman}, employs a unified token mixer to combine sliding local attention~\cite{hassani2023neighborhood} and SS2D~\cite{liu2024vmamba}, achieving competitive performance and a favorable tradeoff in comparison to leading ViTs. However, since these efforts represent early attempts to integrate Transformers and Mamba for vision tasks, attention- and SSM-based modules are simply stacked in their token mixers. There remains a lack of effective methods to achieve a deeper integration between Transformer and Mamba layers, thereby giving rise to a powerful vision backbone that can surpass leading ViTs in terms of both efficiency and performance.
\par
% \begin{itemize}
% To better encapsulate multi-scale clues and long-range dependencies modeling capability into a hybrid token mixer
% , we propose a . MASS differs from our previously proposed LASS in two ways:
In this work, we propose a novel hybrid token mixer, termed \textbf{M}ulti-scale \textbf{A}ttention-augmented \textbf{S}tate \textbf{S}pace {M}odel (MASS), which takes advantage of the strengths of both self-attention and SSM. Specifically, we first introduce an adaptive multi-scale attention (AMA) mechanism, comprising two complementary pathways: (1) regular sliding local attention (SLA) that captures fine-grained spatial details; and (2) dilated sliding attention (DLA) that adaptively adjusts dilation rates to model long-range dependencies. The motivation behind this design is encouraging feature and context representation at multiple granularities. The attention matrices in this mechanism possess dynamic spatial dependencies at multiple scales. Second, to achieve a deeper integration between SSM and self-attention layers, the hidden states of the SSM interact with the aforementioned multi-scale attention matrices via a variant of cross-attention. This design aims to dynamically enhance two-dimensional spatial dependencies and alleviate causality introduced by sequential scanning, thereby improving the spatial perception and dynamic modeling capabilities of SSM. Overall, our MASS effectively encapsulates adaptive multi-scale representation and long-range dependency modeling into a hybrid token mixer.
\par
By hierarchically stacking the MASS token mixer and a feedforward network (FFN) layer, we propose a versatile Transformer-Mamba hybrid vision backbone architecture termed A2Mamba. As shown in Fig.~\ref{fig:acc_plot}, A2Mamba demonstrates remarkably better performance than advanced ConvNets, Transformers and Mamba-based architectures on diverse vision tasks. For instance, our A2Mamba-S model, with approximately 30M parameters only, achieves an impressive top-1 accuracy of 84.7\%, surpassing RMT-S~\cite{fan2023rmt} and TransNeXt-T~\cite{shi2023transnext} by 0.6\% and 0.7\%, respectively, while having higher efficiency. Moreover, A2Mamba-S even outperforms hybrid MambaVision-B~\cite{hatamizadeh2024mambavision} by 0.5\% in top-1 accuracy with only about one-third of the computational complexity. A2Mamba consistently exhibits superior performance over other baselines in dense prediction tasks. For example, in the task of semantic segmentation with UperNet~\cite{xiao2018unified}, A2Mamba-B outperforms BiFormer-B \cite{zhu2023biformer} and UniFormer-B~\cite{li2022uniformer} by 2.3\% and 3.3\% in mIoU, respectively. Meanwhile, in the task of object detection and instance segmentation with Cascade Mask R-CNN~\cite{cai2019cascade}, A2Mamba-L leads CAFormer-M36 \cite{metaformer2024} and MogaNet-L \cite{li2023moganet} by 1.8\%/1.6\% and 2.3\%/2.0\% in AP$^b$/AP$^m$, respectively. These experimental results demonstrate that A2Mamba possesses stronger global modeling and local detail preservation capabilities.
\par
A preliminary version of this work has been published in CVPR 2025~\cite{fu2025segman}. In the preliminary version, our contributions are summarized as follows.
\begin{enumerate}
\item We introduce a novel vision backbone architecture termed SegMAN Encoder featuring a hybrid LASS mixer. LASS synergistically combines \textbf{L}ocal \textbf{A}ttention with \textbf{S}tate \textbf{S}pace Models for both efficient local detail encoding and global context modeling.
\item We propose Mamba-based Multi-Scale Context Extraction (MMSCopE), a novel feature decoder specifically designed for semantic segmentation tasks. MMSCopE operates on multi-scale feature maps that adaptively scale with the input resolution, surpassing previous approaches in both fine-grained detail preservation and omni-scale context modeling.
\item A strong segmentation network architecture, SegMAN, is devised by integrating SegMAN Encoder and MMSCopE. Extensive experiments on semantic segmentation tasks demonstrate the superior performance and competitive efficiency of our method.
\end{enumerate}
\par
In this extended version, we aim to further unleash the potential of Transformer-Mamba hybrid architectures for visual recognition. Compared to our conference paper, this version presents substantial improvements in the following aspects.

\begin{enumerate}
\item We propose a new \textbf{hybrid token mixer} termed MASS, which can more deeply integrate self-attention and SSM, enabling strong multi-scale context modeling and long-range dependency modeling capabilities within a single mixer. Note that the MASS token mixer is a more powerful evolution of the LASS token mixer in the conference paper.

\item Building upon MASS, we propose a stronger \textbf{vision backbone architecture} termed A2Mamba, which encodes more discriminative feature representations for various visual recognition tasks. Furthermore, we leverage MASS to construct a new decoder for semantic segmentation, dubbed MASS-based multi-scale refinement (MM-Refine) module, which is combined with A2Mamba to form a new segmentation network architecture, SegMAN-V2.

\item We have conducted more extensive experimental validations of our architectures on a broader range of visual recognition tasks, including image classification under diverse resolutions and dense predictions including semantic segmentation, object detection, and instance segmentation. Extensive results demonstrate that our method outperforms all existing baselines while incurring comparable computational costs.
\end{enumerate}

% On the other hand, we have conducted more extensive experimental evaluations, including image classification under diverse resolutions, dense predictions including semantic segmentation, and object detection and instance segmentation.
% \textbf{More powerful hybrid token mixer}.
% \textbf{More robust vision foundation network}.
% \textbf{More significant performance gains}.

\section{Related Works}
\subsection{ConvNets}
Since the advent of AlexNet \cite{krizhevsky2012imagenet}, ConvNets have unleashed the potential of deep learning and have gradually become the mainstream architecture for visual recognition. Initially, ConvNet designs focused on employing small kernels (i.e., 3$\times$3) to construct deep networks, gradually increasing the receptive field, such as VGGNet \cite{simonyan2014VGG}, ResNet \cite{he2016deep}, and DenseNet \cite{huang2017densely}. However, modern ConvNet designs \cite{liu2022convnet,woo2023convnext,liu2022more,ding2023unireplknet,chen2024pelk}, exemplified by ConvNeXt \cite{liu2022convnet}, have shifted the focus towards increasing kernel sizes to enlarge the receptive field more quickly, aiming to achieve comparable performance with Transformer- and Mamba-based models. Meanwhile, gating mechanisms have been successfully integrated with modern ConvNets to boost performance \cite{rao2022hornet,yang2022focalnet,li2023moganet}. More recently, OverLoCK \cite{lou2025overlock} has reinvented ConvNet architecture by drawing inspiration from biological top-down neural attention \cite{saalmann2007neural}, significantly outperforming previous ConvNets on various vision tasks. However, it remains challenging to simultaneously obtain more informative multi-scale representations and global dependencies across network layers, which this paper aims to explore a more robust solution.

\subsection{Vision Transformers}
The emergence of ViT \cite{dosovitskiy2020image,Shi_2026_CVPR} has inspired the exploration of multi-head self-attention (MHSA) in the visual domain, with many subsequent works building vision backbone models centered around MHSA. However, vanilla MHSA suffers from quadratic complexity, leading to high computational costs in long-sequence modeling. To this end, various efficient attention mechanisms have been proposed to capture long-range dependencies while maintaining computational efficiency, such as window attention \cite{liu2021swin,dong2022cswin,hassani2023neighborhood,zhu2023biformer}, spatial-reduction attention \cite{wang2021pyramid,wang2021pvtv2}, and dilated attention \cite{tu2022maxvit,wang2023crossformer++}. To further boost performance, BiFormer \cite{zhu2023biformer} introduces bi-level routing attention that captures local-range dependencies in a coarse-to-fine approach. Recently, RMT \cite{fan2023rmt} proposed Manhattan attention, which injects a spatial prior into attention calculation for more accurate global information perception. Despite achieving notable results, the efficient attention mechanisms used in these works generally sacrifice sequence length to progressively capture long-range contexts. In contrast, this paper aims to develop a hybrid architecture that combines multi-scale attention and State Space Models (SSM) \cite{gu2023mamba} to model both fine-grained multi-scale clues and global contexts without reducing sequence length, resulting in a stronger vision architecture.
\begin{figure*}[t]
\centering
\includegraphics[width=0.75\textwidth]{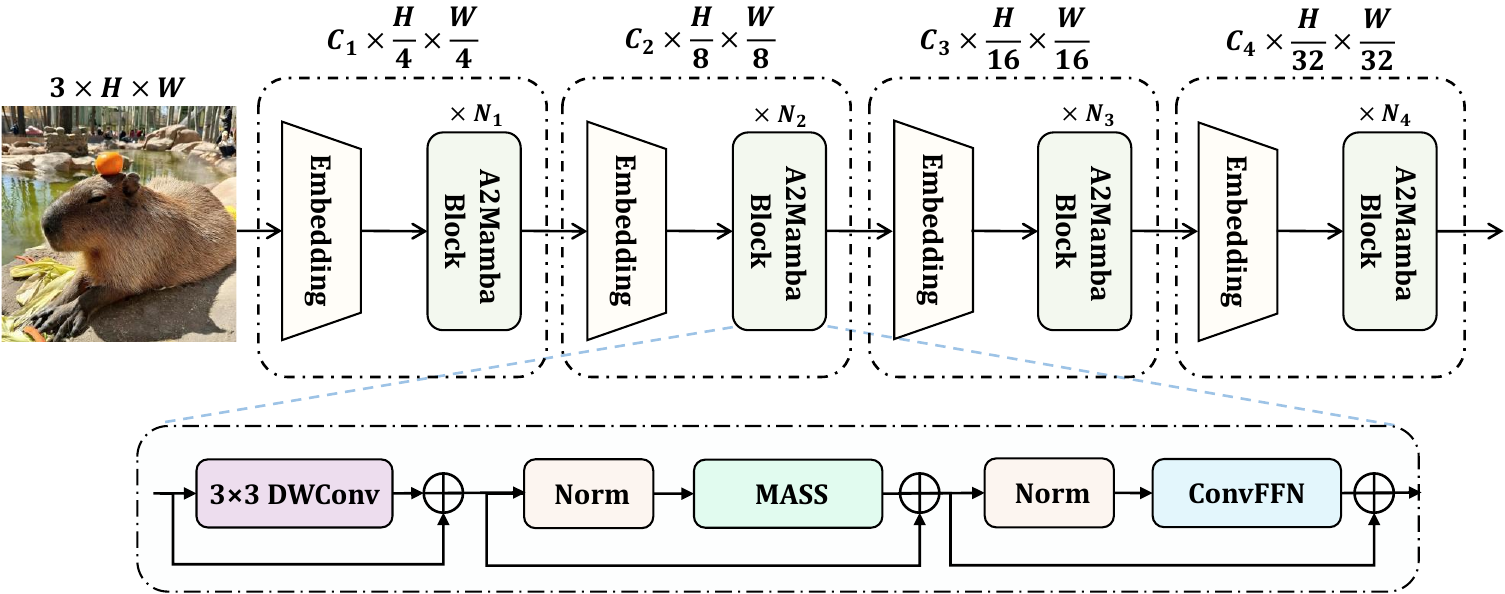}
\caption{The overall architecture of the proposed A2Mamba.}
\label{fig:net}
\end{figure*}
\subsection{Vision Mamba}
Inspired by the outstanding performance of Mamba~\cite{gu2023mamba} in Natural Language Processing (NLP) tasks, researchers have extended its application to computer vision tasks. As the core of Mamba, State Space Models (SSM) can model long-range dependencies with linear-time complexity, demonstrating excellent performance in vision tasks. ViM \cite{zhu2024vision} first introduces a bidirectional SSM module and constructs a plain architecture similar to ViT \cite{dosovitskiy2020image}. VMamba \cite{liu2024vmamba} extends the scanning order to include four directions and presents an early SSM-based hierarchical architecture. Subsequently, a series of representative Mamba-based vision backbone models have been proposed \cite{huang2024localmamba,yang2024plainmamba,pei2024efficientvmamba,xiao2024spatialmamba,lou2025sparx}. For instance, Spatial-Mamba \cite{xiao2024spatialmamba} proposes a structured SSM to enhance the spatial perception of image structure. SparX-Mamba \cite{lou2025sparx} focuses on improving the architecture of Mamba-based networks by proposing a new sparse skip-connection mechanism. This work employs multi-scale self-attention to inherently and dynamically enhance the representational ability of SSM, thereby further unleashing the potential of Mamba-based models in vision tasks.

\subsection{Hybrid Vision Backbone Architectures}
Hybrid vision models have emerged as a promising direction in visual recognition. Previously, various Transformer-ConvNet hybrid models have been extensively studied, showcasing excellent performance \cite{guo2022cmt,pan2022integration,chen2022mixformer,tu2022maxvit,lou2023transxnet,li2022uniformer,metaformer2024}. The primary advantage of hybrid vision models lies in the ability to leverage the strengths of both sub-mixers, such as ACmix \cite{pan2022integration} and MixFormer \cite{chen2022mixformer}, which parallel depthwise convolution (DWConv) and shifted window attention. Recently, TransNeXt \cite{shi2023transnext} presents a foveal self-attention mechanism and ConvGLU, developing a powerful Transformer-ConvNet hybrid vision backbone architecture that demonstrated notable results on various vision tasks. Since the introduction of Mamba, integrating Mamba into hybrid models has shown promising performance. MambaVision \cite{hatamizadeh2024mambavision} integrates Conv, SSM, and MHSA into a single network, although demonstrating high efficiency, its performance, however, still lags behind advanced vision backbone architectures. Our preliminary work, SegMAN~\cite{fu2025segman}, proposes an effective Transformer-Mamba hybrid vision backbone and an accompanying Mamba-based decoder, demonstrating compelling performance improvements over other baselines in semantic segmentation tasks. In this work, we further unleash the potential of Transformer-Mamba hybrid vision architectures by introducing a new and more powerful token mixer termed multi-scale attention-augmented SSM, which more deeply integrate attention with state space models.

\begin{figure*}[t]
\centering
\includegraphics[width=0.725\textwidth]{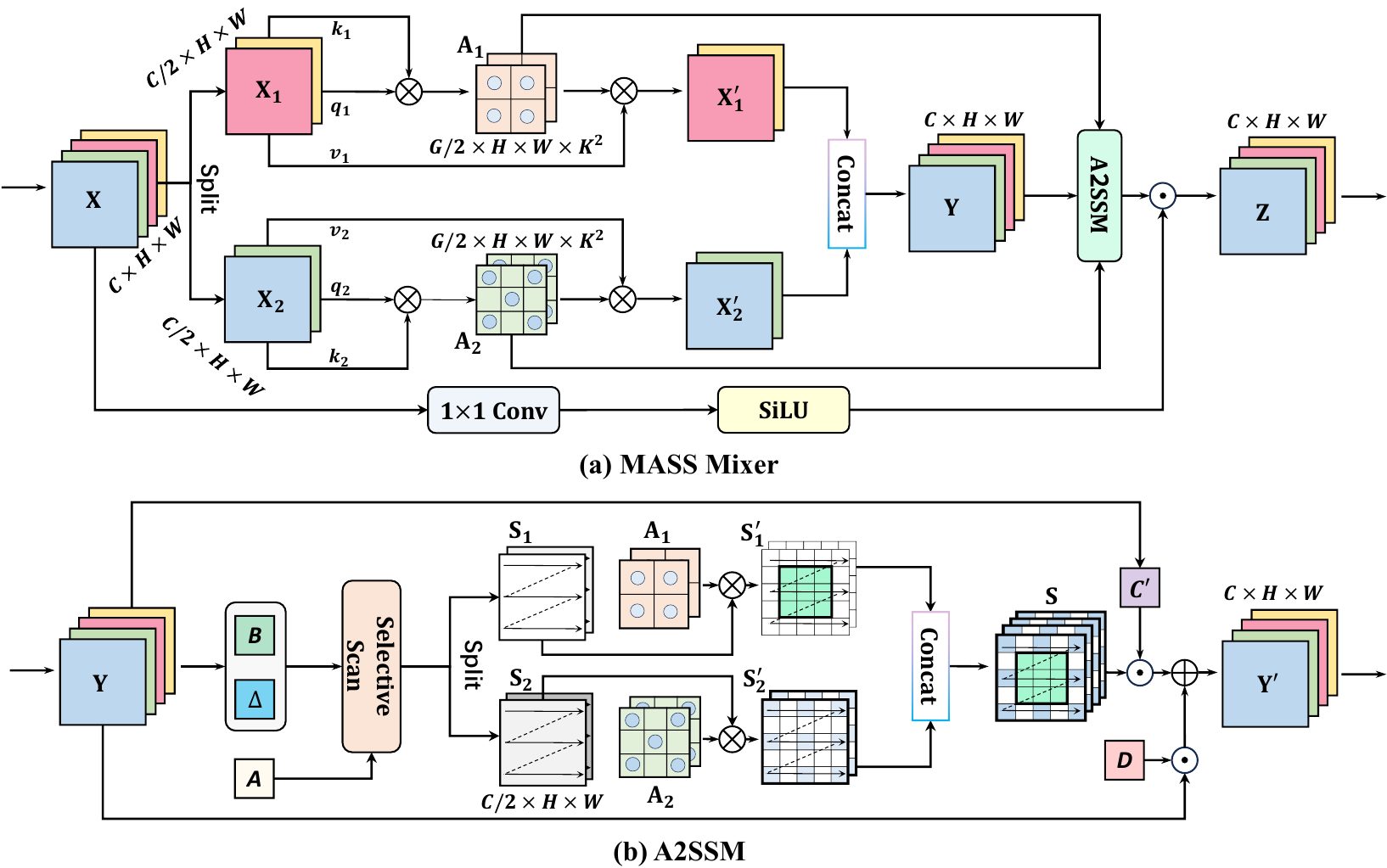}
\caption{Workflow of our MASS token mixer.}
\label{fig:mixer}
\end{figure*}

\section{Method}
In this section, we first briefly review the network architecture in our preliminary work~\cite{fu2025segman}. Then, we elaborate an upgraded version with remarkable performance improvements.
\subsection{A Recap of SegMAN}
Our earlier work in \cite{fu2025segman} represents the early attempt to explore the combination of local self-attention and state space models to build a strong vision backbone architecture, i.e., SegMAN Encoder. The token mixer consists of two complementary stacked modules: Sliding Local Attention (SLA)~\cite{hassani2023neighborhood} for capturing local details and selective scan 2D (SS2D)~\cite{liu2024vmamba} for modeling long-range dependencies. Unlike previous works that model long-range dependencies using spatially subsampled self-attention to reduce sequence length, the linear-time complexity of recent state space models enables our SegMAN Encoder to model global information without sacrificing sequence length, allowing for the preservation of fine-grained spatial information, which is crucial for dense predictions. In the ImageNet-1K classification task, SegMAN Encoder demonstrates excellent performance, significantly outperforming previous ConvNets, Transformers, and Mamba-based architectures, while being on par with advanced Transformer-based architectures, i.e., RMT~\cite{fan2023rmt} and TransNeXt~\cite{shi2023transnext}.
\par
On the other hand, we also propose a Mamba-based decoder, which incorporates a novel Mamba-based multi-scale context extraction (MMSCopE) module, for semantic segmentation. In practice, MMSCopE first computes features on multiple scales and then feeds them into SS2D. The motivation behind this design is that multi-scale features can promote context modeling at various granularities, leading to better semantic segmentation results. By integrating the proposed encoder and decoder, we introduce a new segmentation network architecture, termed SegMAN, which is evaluated on three challenging datasets, including ADE20K~\cite{zhou2017scene}, Cityscapes~\cite{cordts2016cityscapes}, and COCO-Stuff~\cite{caesar2018cocostuff}, outperforming previous state-of-the-art segmentation network architectures such as SegNeXt~\cite{guo2022segnext} and VWFormer~\cite{yan2024vwformer} by a significant margin.

\subsection{Overall Architecture of A2Mamba}
In this work, we propose a novel hybrid vision backbone architecture, A2Mamba, which takes advantage of the strengths of both Transformer and Mamba architectures. A2Mamba is a comprehensively upgraded version of SegMAN Encoder, offering significant improvements in both performance and efficiency. As shown in Fig.~\ref{fig:net}, A2Mamba is a pyramid architecture with four stages, as in previous work~\cite{he2016deep,liu2021swin,liu2022convnet,wang2021pvtv2}. The downsampling factor in each stage is $\frac{1}{4}$, $\frac{1}{8}$, $\frac{1}{16}$, and $\frac{1}{32}$, respectively, while the channel dimension increases with depth. For classification tasks, the output of the deepest stage is fed into a classifier to generate image-level predictions. In contrast, hierarchical features are employed for dense prediction tasks, such as object detection and semantic segmentation.
\par
The key layers of A2Mamba are A2Mamba Blocks, each of which is primarily composed of three components: a residual 3$\times$3 Depthwise Convolution (DWConv) that enhances positional information, a novel \textbf{M}ulti-scale \textbf{A}ttention-augmented \textbf{S}tate \textbf{S}pace Model (MASS) that serves as a core token mixer to capture omni-scale contextual information, and a Convolutional Feedforward Network (ConvFFN)~\cite{wang2021pvtv2} that boosts channel diversity.

\begin{figure*}[t]
\centering
\includegraphics[width=0.9\textwidth]{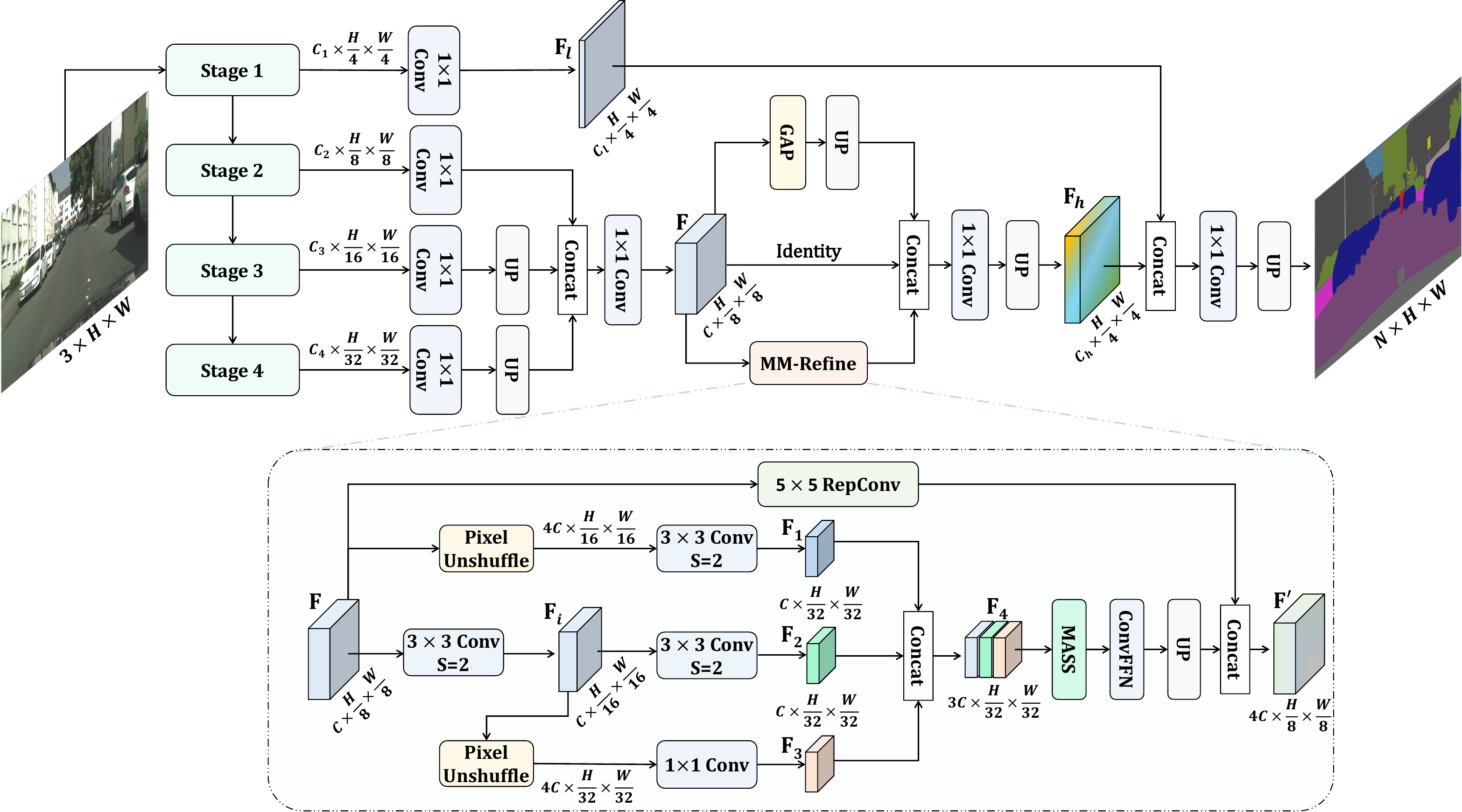}
\caption{The overall architecture of SegMAN-V2.}
\label{fig:decoder}
\end{figure*}
\subsection{MASS Token Mixer}
\label{sec:mixer}
\textbf{Adaptive Multi-scale Attention}. 
% Departing from the LASS mixer in the SegMAN Encoder, 
The proposed MASS enhances its contextual modeling capability by integrating dynamic multi-scale aggregation with long-range propagation, while using a gating mechanism~\cite{gu2023mamba,li2023moganet} to further eliminate contextual noise. As illustrated in Fig.~\ref{fig:mixer} (a), given an input feature map $\mathbf{X} \in \mathbb{R}^{C \times H \times W}$ where $C$ denotes the channel dimension and $H \times W$ the spatial dimensions, we first evenly partition $\mathbf{X}$ channel-wise into $\{ \mathbf{X}_1, \mathbf{X}_2 \} \in \mathbb{R}^{C/2 \times H \times W}$. Then, $\mathbf{X}_1$ is processed with standard SLA~\cite{hassani2023neighborhood}. Specifically, multi-head self-attention (MHSA)~\cite{vaswani2017attention} is computed on $\mathbf{X}_1$ within a sliding window where the only query is located at the center, generating an attention map $\mathbf{A}_1 \in \mathbb{R}^{G/2 \times H \times W \times K^{2}}$, where $G$ is the number of attention heads over the original $\mathbf{X}$ and $K^{2}$ denotes the window size. The attention map dynamically aggregates fine-grained local neighborhoods in $\mathbf{X}_1$ through attention-weighted summation to produce a new feature map $\mathbf{X}'_1$. Meanwhile, $\mathbf{X}_2$ is processed with dilated local attention (DLA)~\cite{hassani2022dilated}, which enlarges receptive fields via a dilation mechanism analogous to dilated convolutions~\cite{chen2017deeplab}. To consistently capture long-range dependencies across different resolutions, the dilation rate $\mathbf{r}$ is determined adaptively as follows, 
\par
\begin{eqnarray}
\mathbf{r} = ({\mathsf{int}} (\frac{H}{K}), {\mathsf{int}} (\frac{W}{K})).
\label{eq:dil}
\end{eqnarray}
\par
The motivation behind this formulation is to make the dilated sliding window have the same size as the input feature map, regardless of the absolute resolution. Thus the scope of attention-based contextual modeling covers the entire input space. Afterwards, the generated feature maps $\{ \mathbf{X}'_1, \mathbf{X}'_2 \}$ are concatenated along the channel dimension to form $\mathbf{Y} \in \mathbb{R}^{C \times H \times W}$. This integration combines fine-grained local details from standard SLA and sparsely sampled long-range dependencies captured by DLA, resulting in an input-dependent multi-scale representation. $\mathbf{Y}$ is fed into an attention-augmented state space model, which will be elaborated below. In practice, we set the window size in the four stages to [11, 9, 7, 7], respectively, following our earlier work~\cite{fu2025segman}.
\par
\textbf{Attention-augmented State Space Model}. 
% Unlike the SegMAN Encoder, which uses the SS2D module to further encode the SLA-based output for global modeling, 
Departing from prior Transformer-Mamba hybrid models that employ SSM or SS2D modules to further encode attention-based outputs for global modeling, we propose a novel attention-augmented state space model (A2SSM) that effectively harnesses pre-computed self-attention maps to boost both spatial perception and dynamic modeling capabilities of SSM. As illustrated in Fig.~\ref{fig:mixer} (b), input $\mathbf{Y}$ is flattened and projected to three input-dependent sequences: $\boldsymbol{\Delta}$, $\textbf{\textit{B}}$, and $\textbf{\textit{C}}'$. Then, $\boldsymbol{\Delta}$, $\textbf{\textit{B}}$, and a learnable vector $\textbf{\textit{A}}$ are used to generate a causal hidden state map (HSM) $\mathbf{S} \in \mathbb{R}^{C \times HW}$ through the selective scan operation, where the $t$-th token dynamically fuses the tokens at all previous positions. Note that {\pcr{d\_state}} is omitted because previous work~\cite{liu2024vmamba,lou2025sparx,xiao2024spatialmamba,fu2025segman} set it to 1 for computational efficiency. 
\par
In vanilla SSM, $\mathbf{S}$ and $\textbf{\textit{C}}'$ are multiplied element-wise to achieve global context modulation. However, our A2SSM can effectively integrate with self-attention to perform more powerful global modeling. We first reshape $\mathbf{S}$ and split it along the channel dimension into $\left \{ \mathbf{S}_1, \mathbf{S}_2 \right \} \in \mathbb{R}^{C/2 \times H\times W}$, and then apply pre-computed attention maps $\left \{ \mathbf{A}_1, \mathbf{A}_2 \right \}$ to them to obtain updated HSMs $\left \{\mathbf{S}'_1, \mathbf{S}'_2 \right \}$. Specifically, $\mathbf{S}_1$ and $\mathbf{S}_2$ are treated as `value' components, whose multi-scale neighborhoods are dynamically aggregated with $\mathbf{A}_1$ and $\mathbf{A}_2$, respectively. The motivation behind this is that $\mathbf{A}_1$ and $\mathbf{A}_2$ capture dynamic affinities in different ranges without causality. In particular, $\mathbf{A}_1$ densely captures the dynamic affinity between every token and its neighbors, while $\mathbf{A}_2$ sparsely captures the dynamic affinity between every token and a set of regularly spaced distant tokens. Meanwhile, both $\mathbf{A}_1$ and $\mathbf{A}_2$ have inherent inductive biases due to their window-shaped spatial scopes. Consequently, by taking $\mathbf{A}_1$ and $\mathbf{A}_2$ into account, the resulting HSMs $\left \{\mathbf{S}'_1, \mathbf{S}'_2 \right \}$ not only have dynamically enhanced spatial coherence and dependency pertaining to a two-dimensional space instead of a one-dimensional sequence, but also suppress causality introduced by sequential scanning in SSM or SS2D. In addition, the inductive biases of our attention maps facilitate the perception of two-dimensional image structures. Therefore, our A2SSM improves the spatial perception and dynamic modeling capacities of vanilla SSM. Next, $\mathbf{S}'_1$ and $\mathbf{S}'_2$ are concatenated along the channel dimension, resulting in $\mathbf{S}' \in \mathbb{R}^{C \times H \times W}$, which is then multiplied element-wise with the reshaped $\textbf{\textit{C}}'$ to achieve enhanced global context modulation. The remaining operations follow vanilla SSM, where a weighted residual connection is added by integrating a learnable weight vector $\textbf{\textit{D}}$ with input $\mathbf{Y}$ before the final output of A2SSM is generated.
\par
Overall, the above design of MASS can be formally expressed as:
\begin{equation}
\label{eq:mass}
\begin{aligned}
& \mathbf{X}_1 ,\mathbf{X}_2 = \mathrm{Split} (\mathbf{X}), \\
& \mathbf{A}_1, \mathbf{X}'_1 = \mathrm{SLA} (\mathbf{X}_1), \\
& \mathbf{A}_2, \mathbf{X}'_2 = \mathrm{DLA} (\mathbf{X}_2), \\
& \mathbf{Y} = \mathrm{Concat} (\mathbf{X}'_1, \mathbf{X}'_2), \\
& \mathbf{Y}' = \mathrm{A2SSM} (\mathbf{Y} , \mathbf{A}_1, \mathbf{A}_2), \\
& \mathbf{Z} = \mathbf{Y}' \odot  \mathrm{SiLU}(\mathrm{Conv_{1\times 1}} (\mathbf{X})). \\
\end{aligned}
\end{equation}
\par
\textbf{Relation to Previous Hybrid Token Mixers}.
\label{sec:hybrid-mixers} 
Although numerous hybrid token mixers have been proposed, existing designs largely fall into two categories.
The first consists of attention-convolution hybrid mixers~\cite{guo2022cmt,tu2022maxvit,shi2023transnext,pan2022integration,lou2023transxnet}, which combine convolution (or local attention) with efficient global attention in parallel or sequential branches.
The second comprises attention-SSM hybrid mixers: LASS~\cite{fu2025segman} sequentially stacks local attention and SS2D, whereas MambaVision~\cite{hatamizadeh2024mambavision} stacks MHSA and SSM blocks in deeper stages.
Despite their different operators, all these designs interact only at the feature level by fusing branch outputs or chaining mixers sequentially, without exploiting the internal and complementary properties of their constituent operators.
Taking the D-Mixer of TransXNet~\cite{lou2023transxnet} as an example, it processes two channel-split feature segments in parallel with a global self-attention branch (OSRA) and an input-dependent depthwise convolution branch (IDConv).
Although MASS also appears to be a dual-branch mixer, the two designs differ fundamentally in the type and location of interaction: D-Mixer fuses the outputs of its two branches by concatenation, whereas MASS additionally reuses the attention maps of AMA inside A2SSM to reorganize the hidden states of the selective scan, achieving an internal integration between attention and SSM that goes beyond feature-level fusion.
Accordingly, the dual-branch structure of AMA, whose branches are both local sliding attention, serves to strengthen multi-scale modeling and constitutes an architectural option rather than the core of our contribution.
To the best of our knowledge, MASS is thus the first hybrid token mixer that achieves such an internal integration between attention and SSM operators.

\subsection{Architecture Variants}
To make more potential applications possible on different devices, our A2Mamba has 5 architectural variants, including Nano (N), Tiny (T), Small (S), Base (B), and Large (L). As listed in Table~\ref{tab:model_variants}, we control the model size by adjusting the number of channels and blocks in each stage. For instance, A2Mamba-S has 4 stages with channel counts $[64, 128, 320, 512]$ and depths $[2, 4, 12, 4]$. The number of attention heads in the four stages is $[2, 4, 10, 16]$, respectively. And the window size used in the four stages is $[11, 9, 7, 7]$, respectively.
\begin{table}[t]
  \centering
  \caption{The configurations of A2Mamba model variants.}
  \resizebox{0.475\textwidth}{!}{
    \begin{tabular}{ccccc}
    \toprule
    {\pcr{A2Mamba}} & {\pcr{Channels}} & {\pcr{Blocks}} & {\pcr{Heads}} & {\pcr{Window Sizes}} \\
    \midrule
    Nano  & [32, 64, 128, 192] & [2, 2, 8, 2] & [2, 2, 4, 8] & [11, 9, 7, 7] \\
    Tiny  & [48, 96, 256, 448] & [2, 2, 10, 2] & [2, 4, 8, 16] & [11, 9, 7, 7] \\
    Small & [64, 128, 320, 512] & [2, 4, 12, 4] & [2, 4, 10, 16] & [11, 9, 7, 7] \\
    Base  & [96, 192, 384, 512] & [4, 6, 12, 6] & [4, 8, 12, 16] & [11, 9, 7, 7] \\
    Large & [112, 224, 512, 720] & [4, 6, 12, 6] & [4, 8, 16, 30] & [11, 9, 7, 7] \\
    \bottomrule
    \end{tabular}%
}
  \label{tab:model_variants}%
\end{table}%
\subsection{SegMAN-V2 for Improved Semantic Segmentation}
\label{sec:seg}
\textbf{Overview}. 
Our preliminary work, SegMAN~\cite{fu2025segman}, primarily focused on semantic segmentation by pairing the SegMAN Encoder with a simple yet efficient decoder centered on MMSCopE. Building upon this, this work not only proposes a new token mixer, MASS, to construct a general-purpose A2Mamba backbone as described above, but also introduces an upgraded segmentation network, termed SegMAN-V2, which assigns A2Mamba as the encoder, while a novel \textbf{M}ASS-based \textbf{M}ulti-scale \textbf{Refine}ment (MM-Refine) module is proposed as the core component of the decoder. As illustrated in Fig.~\ref{fig:decoder}, our decoder aggregates features at multiple levels of abstraction in A2Mamba (i.e. from low-level features in stage 1 to high-level features in stage 4), as in previous work~\cite{xie2021segformer, guo2022segnext}. Specifically, we employ three parallel 1$\times$1 convolution layers to project feature maps in stages $\left \{ 2,3,4 \right \}$ to a lower dimension. Then, we upsample projected feature maps from stages 3 and 4 using bilinear interpolation to match the spatial dimensions of the feature map projected from stage 2. The three transformed feature maps are concatenated and passed through another 1$\times$1 convolution layer, yielding a fused feature map $\mathbf{F} \in \mathbb{R}^{C \times \frac{H}{8} \times \frac{W}{8}}$. Next, $\mathbf{F}$ is further encoded by multiple operators, including global average pooling (GAP) to obtain the image-level global context, an identity mapping to retain the original information and smooth training, and a new MM-Refine module to capture rich multi-scale contextual information. The outputs of these operators are concatenated and subsequently fed into a linear layer followed by a bilinear interpolation layer, resulting in a feature map $\mathbf{F}_h \in \mathbb{R}^{C \times \frac{H}{4} \times \frac{W}{4}}$. Afterwards, we perform low-level enhancement to further refine spatial details~\cite{chen2018encoder,yan2024vwformer}. Namely, the output of stage 1 in A2Mamba is linearly projected into a lower-dimensional feature space $\mathbf{F}_l \in \mathbb{R}^{C_l \times \frac{H}{4} \times \frac{W}{4}}$, which is concatenated with $\mathbf{F}_h$, and fed into a 1$\times$1 convolution layer to fuse together low-level spatial details and high-level contextual information. Finally, the fused feature map is upsampled to produce dense segmentation predictions. By integrating A2Mamba and this decoder, we obtain an upgraded network architecture for semantic segmentation, termed SegMAN-V2.
\par
% The main differences between SegMAN-V2 and the original version are:
% \begin{itemize}
%     \item A more powerful vision backbone is used as the feature encoder, with the SegMAN Encoder evolving into the proposed A2Mamba.
%     \item A new MM-Refine is introduced to replace the original MMSCopE feature extractor in the feature decoder.
%     \item Low-level features are utilized to enhance boundary capture in the segmentation network.
% \end{itemize}
\begin{table*}[!t]
  \centering
  \caption{A comprehensive comparison of image classification on ImageNet-1K with 224$\times$224 inputs. \#F and \#P denote the FLOPs and number of Params of a model, respectively. Type refers to model type, where ``C", ``T", ``M", and ``H" refer to ConvNet, Transformer, Mamba, and hybrid models, respectively.}
  \resizebox{0.9\textwidth}{!}{
    \begin{tabular}{lccccrlcccc}
\cmidrule{1-5}\cmidrule{7-11}    Method & Type  & \# P (M) & \# F (G) & Acc. (\%) &       & Method & Type  & \# P (M) & \# F (G) & Acc. (\%) \\
\cmidrule{1-5}\cmidrule{7-11}    PVTv2-B0\cite{wang2021pvtv2} & T     & 4     & 0.6   & 70.5  &       & Swin-S\cite{liu2021swin} & T     & 50    & 8.7   & 83.0  \\
    QuadMamba-Li\cite{xie2024quadmamba} & M     & 5     & 0.8   & 74.2  &       & ConvNeXt-S\cite{liu2022convnet} & C     & 50    & 8.7   & 83.1  \\
    MSCAN-T\cite{guo2022segnext} & C     & 4     & 0.9   & 75.9  &       & MambaVision-S\cite{hatamizadeh2024mambavision} & H     & 50    & 7.5   & 83.3  \\
    ConvNeXt-V2-A\cite{woo2023convnext} & C     & 4     & 0.5   & 76.2  &       & FocalNet-S\cite{yang2022focalnet} & C     & 50    & 8.7   & 83.5  \\
    EfficientVMamba-T\cite{pei2024efficientvmamba} & M     & 6     & 0.8   & 76.5  &       & InceptionNeXt-S\cite{yu2023inceptionnext} & C     & 49    & 8.4   & 83.5  \\
    UniRepLKNet-A\cite{ding2023unireplknet} & C     & 4     & 0.6   & 77.0  &       & PVTv2-B4\cite{wang2021pvtv2} & T     & 63    & 10.1  & 83.6  \\
    MSVMamba-N\cite{shi2024msvmamba} & M     & 7     & 0.9   & 77.3  &       & VMamba-S\cite{liu2024vmamba} & M     & 50    & 8.7   & 83.6  \\
    \rowcolor[rgb]{ .867,  .922,  .969} \textbf{SegMAN-T Encoder}\cite{fu2025segman} & H     & 4     & 0.7   & 76.2  & \cellcolor[rgb]{ 1,  1,  1} & \cellcolor[rgb]{ 1,  1,  1}NAT-S\cite{hassani2023neighborhood} & \cellcolor[rgb]{ 1,  1,  1}T & \cellcolor[rgb]{ 1,  1,  1}51  & \cellcolor[rgb]{ 1,  1,  1}7.8  & \cellcolor[rgb]{ 1,  1,  1}83.7  \\
    \rowcolor[rgb]{ 0.8, 1, 1} \textbf{A2Mamba-N} & H     & 4     & 0.8   & $\mathbf{78.7}$ & \cellcolor[rgb]{ 1,  1,  1} & \cellcolor[rgb]{ 1,  1,  1}LocalVMamba-S\cite{huang2024localmamba} & \cellcolor[rgb]{ 1,  1,  1}M & \cellcolor[rgb]{ 1,  1,  1}50  & \cellcolor[rgb]{ 1,  1,  1}11.4  & \cellcolor[rgb]{ 1,  1,  1}83.7  \\
\cmidrule{1-5}    PVTv2-B1\cite{wang2021pvtv2} & T     & 14    & 2.1   & 78.7  &       & RDNet-S\cite{kim2024densenets} & C     & 50    & 8.7   & 83.7  \\
    EfficientVMamba-S\cite{pei2024efficientvmamba}  & M     & 11    & 1.3   & 78.7  &       & QuadMamba-B\cite{xie2024quadmamba} & M     & 50    & 9.3   & 83.8  \\
    MSVMamba-M\cite{shi2024msvmamba} & M     & 12    & 1.5   & 79.8  &       & SLaK-S\cite{liu2022more} & C     & 55    & 9.8   & 83.8  \\
    RegionViT-T\cite{chen2021regionvit} & T     & 14    & 2.4   & 80.4  &       & UniFormer-B\cite{li2022uniformer} & H     & 50    & 8.3   & 83.9  \\
    MPViT-XS\cite{lee2022mpvit} & T     & 11    & 2.9   & 80.9  &       & PeLK-S\cite{chen2024pelk} & C     & 50    & 10.7  & 83.9  \\
    ConvNeXt-V2-N\cite{woo2023convnext} & C     & 16    & 2.5   & 81.2  &       & UniRepLKNet-S\cite{ding2023unireplknet} & C     & 56    & 9.1   & 83.9  \\
    BiFormer-T\cite{zhu2023biformer} & T     & 13    & 2.2   & 81.4  &       & HorNet-S\cite{rao2022hornet} & C     & 50    & 8.8   & 84.0  \\
    Conv2Former-N\cite{HouConv2Former} & C     & 15    & 2.2   & 81.5  &       & MSVMamba-S\cite{shi2024msvmamba} & M     & 50    & 8.8   & 84.1  \\
    UniRepLKNet-N\cite{ding2023unireplknet}  & C     & 18    & 2.8   & 81.6  &       & MambaOut-S\cite{yu2024mambaout} & C     & 48    & 9.0   & 84.1  \\
    NAT-M\cite{hassani2023neighborhood} & T     & 20    & 2.7   & 81.8  &       & Conv2Former-S\cite{HouConv2Former} & C     & 50    & 8.7   & 84.1  \\
    SMT-T\cite{lin2023smt} & H     & 12    & 2.4   & 82.2  &       & InternImage-S\cite{wang2022internimage} & C     & 50    & 8.0   & 84.2  \\
    RMT-T\cite{fan2023rmt} & T     & 14    & 2.7   & 82.4  &       & SparX-Mamba-S\cite{lou2025sparx} & M     & 47    & 9.3   & 84.2  \\
    TransNeXt-M\cite{shi2023transnext} & T     & 13    & 2.7   & 82.5  &       & BiFormer-B\cite{zhu2023biformer} & T     & 57    & 9.8   & 84.3  \\
    \rowcolor[rgb]{ 0.8, 1, 1} \textbf{A2Mamba-T} & H     & 15    & 2.7   & $\mathbf{83.0}$ & \cellcolor[rgb]{ 1,  1,  1} & \cellcolor[rgb]{ 1,  1,  1}MogaNet-B\cite{li2023moganet} & \cellcolor[rgb]{ 1,  1,  1}C & \cellcolor[rgb]{ 1,  1,  1}44  & \cellcolor[rgb]{ 1,  1,  1}9.9  & \cellcolor[rgb]{ 1,  1,  1}84.3  \\
\cmidrule{1-5}    Swin-T\cite{liu2021swin} & T     & 28    & 4.5   & 81.3  &       & MLLA-S\cite{han2025mlla} & T     & 43    & 7.3   & 84.4  \\
    EfficientVMamba-B\cite{pei2024efficientvmamba} & M     & 33    & 4.0   & 81.8  &       & MaxViT-S\cite{tu2022maxvit} & H     & 69    & 11.7  & 84.5  \\
    PVTv2-B2\cite{wang2021pvtv2} & T     & 25    & 4.0   & 82.0  &       & CAFormer-M36\cite{metaformer2024} & H     & 57    & 12.8  & 84.5  \\
    ConvNeXt-T\cite{liu2022convnet}  & C     & 29    & 4.5   & 82.1  &       & Spatial-Mamba-S\cite{xiao2024spatialmamba} & M     & 43    & 7.1   & 84.6  \\
    FocalNet-T\cite{yang2022focalnet} & C     & 29    & 4.5   & 82.3  &       & TransNeXt-S\cite{shi2023transnext} & T     & 50    & 10.3  & 84.7  \\
    InceptionNeXt-T\cite{yu2023inceptionnext} & C     & 28    & 4.2   & 82.3  &       & RMT-B\cite{fan2023rmt} & T     & 54    & 10.4  & 85.0  \\
    QuadMamba-S\cite{xie2024quadmamba} & M     & 31    & 5.5   & 82.4  &       & \cellcolor[rgb]{ .867,  .922,  .969}\textbf{SegMAN-B Encoder}\cite{fu2025segman} & \cellcolor[rgb]{ .867,  .922,  .969}H & \cellcolor[rgb]{ .867,  .922,  .969}45  & \cellcolor[rgb]{ .867,  .922,  .969}9.9  & \cellcolor[rgb]{ .867,  .922,  .969}85.1  \\
    ConvNeXt-V2-T\cite{woo2023convnext} & C     & 29    & 4.5   & 82.5  &       & \cellcolor[rgb]{ 0.8, 1, 1}\textbf{A2Mamba-B} & \cellcolor[rgb]{ 0.8, 1, 1}H & \cellcolor[rgb]{ 0.8, 1, 1}51  & \cellcolor[rgb]{ 0.8, 1, 1}10.7  & \cellcolor[rgb]{ 0.8, 1, 1}$\mathbf{85.7}$ \\
\cmidrule{7-11}    SLaK-T\cite{liu2022more}& C     & 30    & 5.0   & 82.5  &       & Swin-B\cite{liu2021swin} & T     & 88    & 15.4  & 83.5  \\
    VMamba-T\cite{liu2024vmamba} & M     & 29    & 4.9   & 82.6  &       & FocalNet-B\cite{yang2022focalnet} & C     & 89    & 15.4  & 83.7  \\
    PeLK-T\cite{chen2024pelk}& C     & 29    & 5.6   & 82.6  &       & PVTv2-B5\cite{wang2021pvtv2} & T     & 82    & 11.8  & 83.8  \\
    CSWin-T\cite{dong2022cswin}& T     & 23    & 4.5   & 82.7  &       & ConvNeXt-B\cite{liu2022convnet} & C     & 89    & 15.4  & 83.8  \\
    LocalVMamba-T\cite{huang2024localmamba} & M     & 26    & 5.7   & 82.7  &       & VMamba-B\cite{liu2024vmamba} & M     & 89    & 15.4  & 83.9  \\
    MambaVision-T2\cite{hatamizadeh2024mambavision} & H     & 35    & 5.1   & 82.7  &       & SLaK-B\cite{liu2022more} & C     & 95    & 17.1  & 84.0  \\
    MambaOut-T\cite{yu2024mambaout} & C     & 27    & 4.5   & 82.7  &       & InceptionNeXt-B\cite{yu2023inceptionnext} & C     & 87    & 14.9  & 84.0  \\
    HorNet-T\cite{rao2022hornet} & C     & 22    & 4.0   & 82.8  &       & CSWin-B\cite{dong2022cswin} & T     & 78    & 15.0  & 84.2  \\
    RDNet-T\cite{kim2024densenets} & C     & 24    & 5.0   & 82.8  &       & MambaVision-B\cite{hatamizadeh2024mambavision} & H     & 98    & 15.0  & 84.2  \\
    UniFormer-S\cite{li2022uniformer} & H     & 22    & 3.6   & 82.9  &       & MambaOut-B\cite{yu2024mambaout} & C     & 85    & 15.8  & 84.2  \\
    MPViT-S\cite{lee2022mpvit} & T     & 23    & 4.7   & 83.0  &       & PeLK-B\cite{chen2024pelk} & C     & 89    & 18.3  & 84.2  \\
    MSVMamba-T\cite{shi2024msvmamba}& M     & 32    & 5.1   & 83.0  &       & ConvNeXt-V2-B\cite{woo2023convnext} & C     & 89    & 15.4  & 84.3  \\
    NAT-T\cite{hassani2023neighborhood} & T     & 28    & 4.3   & 83.2  &       & MPViT-B\cite{lee2022mpvit} & T     & 75    & 16.4  & 84.3  \\
    Conv2Former-T\cite{HouConv2Former} & C     & 27    & 4.4   & 83.2  &       & NAT-B\cite{hassani2023neighborhood} & T     & 90    & 13.7  & 84.3  \\
    UniRepLKNet-T\cite{ding2023unireplknet} & C     & 31    & 4.9   & 83.2  &       & HorNet-B\cite{rao2022hornet} & C     & 87    & 15.6  & 84.3  \\
    MogaNet-S\cite{li2023moganet}& C     & 25    & 5.0   & 83.4  &       & MSVMamba-B\cite{shi2024msvmamba} & M     & 91    & 16.3  & 84.4  \\
    CMT-S\cite{guo2022cmt} & T     & 25    & 4.0   & 83.5  &       & RDNet-B\cite{kim2024densenets} & C     & 87    & 15.4  & 84.4  \\
    MLLA-T\cite{han2025mlla} & T     & 25    & 4.2   & 83.5  &       & Conv2Former-B\cite{HouConv2Former} & C     & 90    & 15.9  & 84.4  \\
    Spatial-Mamba-T\cite{xiao2024spatialmamba}& M     & 27    & 4.5   & 83.5  &       & SparX-Mamba-B\cite{lou2025sparx} & M     & 84    & 15.9  & 84.5  \\
    SparX-Mamba-T\cite{lou2025sparx}& M     & 27    & 5.2   & 83.5  &       & MogaNet-L\cite{li2023moganet} & C     & 83    & 15.9  & 84.7  \\
    InternImage-T\cite{wang2022internimage}& C     & 30    & 5.0   & 83.5  &       & TransNeXt-B\cite{shi2023transnext} & T     & 90    & 18.4  & 84.8  \\
    CAFormer-S18\cite{metaformer2024}& H     & 26    & 4.1   & 83.6  &       & MaxViT-B\cite{tu2022maxvit} & H     & 120   & 24.0  & 84.9  \\
    MaxViT-T\cite{tu2022maxvit} & H     & 31    & 5.6   & 83.7  &       & InternImage-B\cite{wang2022internimage} & C     & 97    & 16.0  & 84.9  \\
    SMT-S\cite{lin2023smt} & H     & 21    & 4.7   & 83.7  &       & MLLA-B\cite{han2025mlla} & T     & 96    & 16.2  & 85.3  \\
    BiFormer-S\cite{zhu2023biformer} & T     & 26    & 4.5   & 83.8  &       & Spatial-Mamba-B\cite{xiao2024spatialmamba} & M     & 95    & 16.8  & 85.3  \\
    TransNeXt-T\cite{shi2023transnext}& T     & 28    & 5.7   & 84.0  &       & CAFormer-B36\cite{metaformer2024} & H     & 99    & 23.2  & 85.5  \\
    RMT-S\cite{fan2023rmt} & T     & 27    & 4.8   & 84.1  &       & RMT-L\cite{fan2023rmt} & T     & 96    & 19.6  & 85.5  \\
    \rowcolor[rgb]{ .867,  .922,  .969} \textbf{SegMAN-S Encoder}\cite{fu2025segman} & H     & 26    & 4.1   & 84.0  & \cellcolor[rgb]{ 1,  1,  1} & \textbf{SegMAN-L Encoder}\cite{fu2025segman} & H     & 81    & 16.8  & 85.5  \\
    \rowcolor[rgb]{ 0.8, 1, 1} \textbf{A2Mamba-S} & H     & 31    & 5.4   & $\mathbf{84.7}$ & \cellcolor[rgb]{ 1,  1,  1} & \textbf{A2Mamba-L} & H     & 95    & 17.4  & $\mathbf{86.1}$ \\
\cmidrule{1-5}\cmidrule{7-11}    \end{tabular}%
}
  \label{tab:cls}%
\end{table*}%
\textbf{MM-Refine}. To encapsulate multi-scale rich contextual information into the above decoder, in this work, we further propose the MM-Refine module, an upgraded version of the MMSCopE module in \cite{fu2025segman}. As shown in Fig.~\ref{fig:decoder}, we improve the downsampling operation in MMSCopE~\cite{fu2025segman} by using fewer parameters while reducing information loss. Specifically, in the first branch, $\mathbf{F}$ is first passed through a pixel unshuffle layer to achieve lossless downsampling, which is then fed into a 3$\times$3 convolution with stride=2 to obtain $\mathbf{F}_1 \in \mathbb{R}^{C \times \frac{H}{32} \times \frac{W}{32}}$. Unlike MMSCopE, which directly uses pixel unshuffle followed by a 1$\times$1 convolution to reduce the resolution to $H/32 \times W/32$, our progressive downsampling approach can better alleviate information loss. In the second branch, we first use a 3$\times$3 convolution with stride=2 to obtain an interim feature $\mathbf{F}_i \in \mathbb{R}^{C \times \frac{H}{16} \times \frac{W}{16}}$, and then an additional 3$\times$3 convolution with stride=2 is used to further reduce the resolution to obtain $\mathbf{F}_2 \in \mathbb{R}^{C \times \frac{H}{32} \times \frac{W}{32}}$. Meanwhile, $\mathbf{F}_i$ is also fed into a pixel unshuffle layer followed by a 1$\times$1 convolution to reduce its resolution to $H/32 \times W/32$, resulting in $\mathbf{F}_3$. The motivation behind this is to efficiently capture multiple regionally aggregated contexts at different scales, that is, $\left \{ \mathbf{F}_1, \mathbf{F}_2, \mathbf{F}_3\right \}$ represent semantic information at multiple granularities. Compared to MMSCopE, MM-Refine's downsampling approach is more progressive and uses fewer convolution layers, resulting in higher efficiency. Finally, $\left \{ \mathbf{F}_1, \mathbf{F}_2, \mathbf{F}_3\right \}$ are concatenated along the channel dimension and fed into the proposed MASS mixer followed by FFN and bilinear upsampling layers. Note that due to a smaller feature resolution, the MASS mixer here adopts global self-attention instead of the multi-scale self-attention used in Section~\ref{sec:mixer}. Since $\left \{ \mathbf{F}_1, \mathbf{F}_2, \mathbf{F}_3\right \}$ encapsulate multi-scale information, MASS can capture rich contextual information for objects with a wide range of sizes.
% In MMSCopE, the input feature $\mathbf{F}$ is first processed through multiple paths to extract multi-scale information. Specifically, $\mathbf{F}$ is passed through pixel unshuffle, followed by a 3$\times$3 convolution with stride=2 and another pixel unshuffle, and then a 5$\times$5 convolution with stride=2 followed by pixel unshuffle. The resulting feature is fused using a linear layer to obtain a low-resolution feature map ($H/32 \times W/32$), which is then fed into a SS2D layer followed by FFN to extract global information. The motivation behind this design is that the features obtained from convolutional downsampling with different kernels can correspond to multi-scale regions in $\mathbf{F}$, allowing SS2D to model global information at different granularities. Note that our previous work has demonstrated that performing SS2D on low-resolution high-channel features can achieve better performance and efficiency compared to high-resolution low-channel features.
% \par

\begin{table}[t]
  \centering
  \caption{A comparison of image classification performance with 384$\times$384 inputs.}
  % \resizebox{0.375\textwidth}{!}{
    \begin{tabular}{lcccc}
    \toprule
    Method & Type  & \# P (M) & \# F (G) & Acc. (\%) \\
    \midrule
    Swin-B\cite{liu2021swin} & T     & 88    & 47    & 84.5  \\
    CSWin-B\cite{dong2022cswin} & T     & 78    & 47    & 85.4 \\
    ConvNeXt-B\cite{liu2022convnet} & C     & 89    & 45    & 85.1  \\
    ConvNeXt-L\cite{liu2022convnet} & C     & 198   & 101   & 85.5  \\
    MaxViT-S\cite{tu2022maxvit} & H     & 69    & 36    & 85.2  \\
    MaxViT-B\cite{tu2022maxvit} & H     & 120   & 74    & 85.7  \\
    TransNeXt-S\cite{shi2023transnext} & T     & 50    & 32    & 86.0  \\
    TransNeXt-B\cite{shi2023transnext} & T     & 90    & 56    & 86.2  \\
    RMT-L\cite{fan2023rmt} & T     & 95    & 59    & 85.5  \\
     \rowcolor[rgb]{ 0.8, 1, 1}\textbf{A2Mamba-B} & H     & 51    & 34    & $\mathbf{86.4}$ \\
     \rowcolor[rgb]{ 0.8, 1, 1}\textbf{A2Mamba-L} & H     & 95    & 54    & $\mathbf{86.7}$ \\
    \bottomrule
    \end{tabular}%
   % }
  \label{tab:cls-384}%
\end{table}%

Despite our careful use of progressive downsampling, certain important local clues may still be lost. To address this, we introduce an additional lightweight convolutional shortcut based on a 5$\times$5 dilated RepConv~\cite{ding2023unireplknet} to strengthen local detail modeling capabilities. The final feature $\mathbf{F}'$ not only possesses rich multi-scale contextual information but also retains local details, both of which are indispensable for high-quality semantic segmentation.
\begin{table}[htbp]
  \centering
  \caption{A comparison of backbone architectures using Mask R-CNN on the COCO dataset. FLOPs are calculated with an image resolution of 800$\times$1280.}
  \resizebox{0.4\textwidth}{!}{
    \begin{tabular}{lcccc}
    \toprule
    Backbone & \# P (M) & \# F (G) & AP$^b$ & AP$^m$ \\
    \midrule
    ConvNeXt-T\cite{liu2022convnet} & 48    & 262   & 46.2  & 41.7  \\
    FocalNet-T\cite{yang2022focalnet} & 49    & 268   & 48.0  & 42.9  \\
    InternImage-T\cite{wang2022internimage} & 49    & 270   & 49.1  & 43.7  \\
    RDNet-T\cite{kim2024densenets} & 43    & 278   & 47.3  & 42.2  \\
    MogaNet-S\cite{li2023moganet} & 45    & 272   & 48.5  & 43.1  \\
    VMamba-T\cite{liu2024vmamba} & 50    & 271   & 48.8  & 43.7  \\
    MSVMamba-T\cite{shi2024msvmamba} & 52    & 275   & 48.7  & 43.4  \\
    Spatial-Mamba-T\cite{xiao2024spatialmamba} & 46    & 261   & 49.3  & 43.6  \\
    SparX-Mamba-T\cite{lou2025sparx} & 47    & 279   & 50.2  & 44.7  \\
    Swin-T\cite{liu2021swin} & 48    & 267   & 46.0  & 41.6  \\
    PVTv2-B2\cite{wang2021pvtv2} & 45    & 309   & 47.8  & 43.1  \\
    CSWin-T\cite{dong2022cswin} & 42    & 279   & 49.0  & 43.6  \\
    MPViT-S\cite{lee2022mpvit} & 43    & 268   & 48.4  & 43.9  \\
    UniFormer-S\cite{li2022uniformer} & 41    & 269   & 48.2  & 43.4  \\
    NAT-T\cite{hassani2023neighborhood} & 48    & 258   & 47.8  & 42.6  \\
    SMT-S\cite{lin2023smt} & 40    & 265   & 49.0  & 43.4  \\
    RMT-S\cite{fan2023rmt} & 45    & 288   & 50.7  & 44.9  \\
    \rowcolor[rgb]{ 0.8, 1, 1} \textbf{A2Mamba-S} & 49    & 283   & $\mathbf{51.5 }$ & $\mathbf{45.3 }$ \\
    \midrule
    ConvNeXt-S\cite{liu2022convnet} & 70    & 348   & 47.9  & 42.9  \\
    FocalNet-S\cite{yang2022focalnet} & 72    & 365   & 49.3  & 43.8  \\
    InternImage-S\cite{wang2022internimage} & 69    & 340   & 49.7  & 44.5  \\
    MogaNet-B\cite{li2023moganet} & 63    & 373   & 50.3  & 44.4  \\
    VMamba-S\cite{liu2024vmamba} & 70    & 384   & 49.9  & 44.2  \\
    MSVMamba-S\cite{shi2024msvmamba} & 70    & 349   & 49.7  & 44.2  \\
    Spatial-Mamba-S\cite{xiao2024spatialmamba} & 63    & 315   & 50.5  & 44.6  \\
    SparX-Mamba-S\cite{lou2025sparx} & 67    & 361   & 51.0  & 45.2  \\
    Swin-S\cite{liu2021swin} & 69    & 354   & 48.2  & 43.2  \\
    PVTv2-B3\cite{wang2021pvtv2} & 65    & 397   & 48.4  & 43.2  \\
    CSWin-S\cite{dong2022cswin} & 54    & 342   & 50.0  & 44.5  \\
    UniFormer-B\cite{li2022uniformer} & 69    & 399   & 50.3  & 44.8  \\
    NAT-S\cite{hassani2023neighborhood} & 70    & 330   & 48.4  & 43.2  \\
    SMT-B\cite{lin2023smt} & 52    & 328   & 49.8  & 44.0  \\
    RMT-B\cite{fan2023rmt} & 73    & 422   & 52.2  & 46.1  \\
    \rowcolor[rgb]{ 0.8, 1, 1} \textbf{A2Mamba-B} & 70    & 410   & $\mathbf{52.7 }$ & $\mathbf{46.8 }$ \\
    \midrule
    ConvNeXt-B\cite{liu2022convnet} & 108   & 486   & 48.5  & 43.5  \\
    FocalNet-B\cite{yang2022focalnet} & 111   & 507   & 49.8  & 44.1  \\
    InternImage-B\cite{wang2022internimage} & 115   & 501   & 50.3  & 44.8  \\
    MogaNet-L\cite{li2023moganet} & 102   & 495   & 50.5  & 44.5  \\
    SparX-Mamba-B\cite{lou2025sparx} & 103   & 498   & 51.8  & 45.8  \\
    Swin-B\cite{liu2021swin} & 107   & 496   & 48.6  & 43.3  \\
    PVTv2-B5\cite{wang2021pvtv2} & 102   & 557   & 48.4  & 42.9  \\
    CSWin-B\cite{dong2022cswin} & 97    & 526   & 50.8  & 44.9  \\
    MPViT-B\cite{lee2022mpvit} & 95    & 503   & 49.5  & 44.5  \\
    \rowcolor[rgb]{ 0.8, 1, 1} \textbf{A2Mamba-L} & 113   & 552   & $\mathbf{53.0 }$ & $\mathbf{46.8 }$ \\
    \bottomrule
    \end{tabular}%
    }
  \label{tab:det1}%
\end{table}%

\begin{table}[htbp]
  \centering
  \caption{A comparison of backbone architectures using Cascade Mask R-CNN on the COCO dataset. FLOPs are calculated with an image resolution of 800$\times$1280.}
  \resizebox{0.4\textwidth}{!}{
    \begin{tabular}{lcccc}
    \toprule
    Backbone & \# P (M) & \# F (G) & AP$^b$ & AP$^m$ \\
    \midrule
    ConvNeXt-T\cite{liu2022convnet} & 86    & 741   & 50.4  & 43.7  \\
    HorNet-T\cite{rao2022hornet} & 80    & 730   & 51.7  & 44.8  \\
    RDNet-T\cite{kim2024densenets} & 81    & 757   & 51.6  & 44.6  \\
    PeLK-T\cite{chen2024pelk} & 86    & 770   & 51.4  & 44.6  \\
    UniRepLKNet-T\cite{ding2023unireplknet} & 89    & 749   & 51.8  & 44.9  \\
    MogaNet-S\cite{li2023moganet} & 78    & 750   & 51.6  & 45.1  \\
    MambaVision-T\cite{hatamizadeh2024mambavision} & 86    & 740   & 51.1  & 44.3  \\
    Swin-T\cite{liu2021swin} & 86    & 745   & 50.4  & 43.7  \\
    PVTv2-B2\cite{wang2021pvtv2} & 83    & 788   & 51.1  & - \\
    CSWin-T\cite{dong2022cswin} & 80    & 757   & 52.5  & 45.3  \\
    % CGViT-T & 85    & 770   & 51.6  & 44.6  \\
    UniFormer-S\cite{li2022uniformer} & 79    & 747   & 52.1  & 45.2  \\
    NAT-T\cite{hassani2023neighborhood} & 85    & 737   & 51.4  & 44.5  \\
    SMT-S\cite{lin2023smt} & 78    & 744   & 51.9  & 44.7  \\
    CAFormer-S18\cite{metaformer2024} & -     & 733   & 51.5  & 44.6  \\
    RMT-S\cite{fan2023rmt} & 83    & 767   & 53.2  & 46.1  \\
    \rowcolor[rgb]{ 0.8, 1, 1} \textbf{A2Mamba-S} & 87    & 762   & $\mathbf{54.0 }$ & $\mathbf{46.6 }$ \\
    \midrule
    ConvNeXt-S\cite{liu2022convnet} & 108   & 827   & 51.9  & 45.0  \\
    HorNet-S\cite{rao2022hornet} & 108   & 827   & 52.7  & 45.6  \\
    RDNet-S\cite{kim2024densenets} & 108   & 832   & 52.3  & 45.3  \\
    PeLK-S\cite{chen2024pelk} & 108   & 874   & 52.2  & 45.3  \\
    UniRepLKNet-S\cite{ding2023unireplknet} & 113   & 835   & 53.0  & 45.9  \\
    MogaNet-B\cite{li2023moganet} & 101   & 851   & 52.6  & 46.0  \\
    MambaVision-S\cite{hatamizadeh2024mambavision} & 106   & 828   & 52.3  & 45.2  \\
    Swin-S\cite{liu2021swin} & 107   & 838   & 51.8  & 44.7  \\
    CSWin-S\cite{dong2022cswin} & 92    & 820   & 53.7  & 46.4  \\
    % GC-ViT-S & 108   & 866   & 52.4  & 45.4  \\
    UniFormer-B\cite{li2022uniformer} & 107   & 878   & 53.8  & 46.4  \\
    NAT-S\cite{hassani2023neighborhood} & 108   & 809   & 52.0  & 44.9  \\
    CAFormer-S36\cite{metaformer2024} & -     & 811   & 53.2  & 46.0  \\
    RMT-B\cite{fan2023rmt} & 111   & 900   & 54.5  & 47.2  \\
    \rowcolor[rgb]{ 0.8, 1, 1} \textbf{A2Mamba-B} & 108   & 889   & $\mathbf{55.4 }$ & $\mathbf{47.6 }$ \\
    \midrule
    ConvNeXt-B\cite{liu2022convnet} & 146   & 964   & 52.7  & 45.6  \\
    HorNet-B\cite{rao2022hornet} & 144   & 969   & 53.3  & 46.1  \\
    RDNet-B\cite{kim2024densenets} & 144   & 971   & 52.3  & 45.3  \\
    PeLK-B\cite{chen2024pelk} & 147   & 1028  & 52.9  & 45.9  \\
    MogaNet-L\cite{li2023moganet} & 149   & 974   & 53.3  & 46.1  \\
    MambaVision-B\cite{hatamizadeh2024mambavision} & 145   & 964   & 52.8  & 45.7  \\
    Swin-B\cite{liu2021swin} & 145   & 982   & 51.9  & 45.0  \\
    CSWin-B\cite{dong2022cswin} & 135   & 1004  & 53.9  & 46.4  \\
    % GC-ViT-B & 146   & 1018  & 52.9  & 45.8  \\
    NAT-B\cite{hassani2023neighborhood} & 147   & 931   & 52.5  & 45.2  \\
    CAFormer-M36\cite{metaformer2024} & -     & 920   & 53.8  & 46.5  \\
    \rowcolor[rgb]{ 0.8, 1, 1} \textbf{A2Mamba-L} & 151   & 1027  & $\mathbf{55.6 } $& $\mathbf{48.1 }$ \\
    \bottomrule
    \end{tabular}%
    }
  \label{tab:det2}%
\end{table}%

\section{Experiments}
\subsection{Image Classification}
\label{sec:cls}
\textbf{Setup.} We evaluate our approach on the ImageNet-1K dataset~\cite{deng2009imagenet} and follow the same experimental setup as previous works \cite{liu2021swin,han2025mlla} to ensure a fair comparison. Specifically, we train all models for 300 epochs using the AdamW optimizer \cite{loshchilov2017decoupled}. The stochastic depth rate \cite{huang2016deep} is set to 0.05, 0.1, 0.2, 0.4, and 0.5 for the A2Mamba-N, -T, -S, -B, and -L models, respectively. After pre-training the base and large models on 224$\times$224 inputs, we further fine-tune them on 384$\times$384 inputs for 30 epochs to evaluate the performance with high-resolution inputs. All experiments are run on 8 NVIDIA H800 GPUs.
\par
\textbf{Results.} As shown in Table \ref{tab:cls}, our previous work, SegMAN Encoder, has already achieved competitive performance with state-of-the-art (SOTA) vision backbone models. However, the upgraded version, A2Mamba, results in significant performance improvement over all previous ConvNet-, Transformer, and Mamba-based models. Specifically, our A2Mamba-S model achieves an impressive top-1 accuracy of 84.7\%, outperforming RMT-S \cite{fan2023rmt} and TransNeXt-T \cite{shi2023transnext} by 0.6\% and 0.7\%, respectively. Furthermore, A2Mamba-B further increases top-1 accuracy to 85.7\%, surpassing MLLA-B \cite{han2025mlla} by 0.4\% while reducing computational complexity by approximately half. Notably, our A2Mamba-L achieves a remarkable 86.1\% top-1 accuracy, outperforming CAFormer-B36 \cite{metaformer2024} by a notable 0.6\% with fewer complexity. As listed in Table \ref{tab:cls-384}, fine-tuning A2Mamba-B on 384$\times$384 inputs yields a top-1 accuracy of 86.4\%, which is better than both TransNeXt-B and RMT-L with only about half the computational complexity. In addition, A2Mamba-L further improves top-1 accuracy to 86.7\%, surpassing its counterparts significantly.

\subsection{Object Detection and Instance Segmentation}
\textbf{Setup}. We evaluate our A2Mamba network architecture on object detection and instance segmentation tasks using the COCO 2017 dataset \cite{lin2014microsoft}. Following the experimental setup of Swin \cite{liu2021swin}, we employ both Mask R-CNN \cite{he2017mask} and Cascade Mask R-CNN \cite{cai2019cascade} frameworks. Our backbone networks are pre-trained on ImageNet-1K and then fine-tuned for 36 epochs with multi-scale training (3$\times$ + MS schedule).
% The training involves 36 epochs with a multi-scale data augmentation 
\par
\textbf{Results}. As shown in Tables~\ref{tab:det1} and \ref{tab:det2}, our model achieves impressive performance on object detection and instance segmentation. For example, with the Mask R-CNN framework, A2Mamba-S outperforms UniFormer-S \cite{li2022uniformer} by a notable margin of 3.3\%/1.9\% in AP$^b$/AP$^m$ and even surpasses CSWin-B by 0.7\%/0.4\% in AP$^b$/AP$^m$ while having only about half the complexity. With the Cascade Mask R-CNN framework, our method exhibits more significant performance gains. For instance, A2Mamba-B surpasses CAFormer-S36 \cite{metaformer2024} by a substantial margin of 2.2\%/1.6\% in AP$^b$/AP$^m$, and it also remarkably outperforms MambaVision-B \cite{hatamizadeh2024mambavision} by 2.6\%/1.9\% in AP$^b$/AP$^m$ while saving about one-third of Params. This notable performance improvement effectively demonstrates the strong capability of our method in modeling multi-scale and global contexts.

\begin{table}[t]
  \centering
  \caption{A comparison of semantic segmentation performance on the ADE20K dataset using various vision backbones with UperNet. FLOPs are calculated for the 512$\times$2048 resolution.}
    \resizebox{0.45\textwidth}{!}{
    \begin{tabular}{l|cccc}
    \toprule
    Backbone & \# P (M) & \# F (G) & mIoU$_{\mathrm{SS}}$ (\%)  & mIoU$_{\mathrm{MS}}$ (\%) \\
    \midrule
    ConvNeXt-T\cite{liu2022convnet}  & 60    & 939   & 46.0  & 46.7  \\
    SLaK-T\cite{liu2022more}& 65    & 936   & 47.6  & - \\
    InternImage-T\cite{wang2022internimage} & 59    & 944   & 47.9  & 48.1  \\
    PeLK-T\cite{chen2024pelk}& 62    & 970   & 48.1  & - \\
    MogaNet-S\cite{li2023moganet} & 55    & 946   & 49.2  & - \\
    % OverLoCK-T & 63    & 969   & 50.3  & 50.8  \\
    VMamba-T\cite{liu2024vmamba} & 62    & 949   & 48.0  & 48.8  \\
    MSVMamba-T\cite{shi2024msvmamba}& 63    & 953   & 47.9  & 48.5  \\
    MambaVision-T\cite{hatamizadeh2024mambavision}& 55    & 945   & 46.0  & - \\
    SparX-Mamba-T\cite{lou2025sparx}& 50    & 954   & 50.0  & 50.8  \\
    Spatial-Mamba-T\cite{xiao2024spatialmamba} & 57    & 936   & 48.6  & 49.4  \\
    CSWin-T\cite{dong2022cswin} & 59    & 959   & 49.3  & 50.7  \\
    UniFormer-S\cite{li2022uniformer} & 52    & 1008  & 47.6  & 48.5  \\
    BiFormer-S\cite{zhu2023biformer} & 55    & 1025  & 49.8  & 50.8  \\
    CAFormer-S18\cite{metaformer2024} & 54    & 1024  & 48.9  & - \\
    TransNeXt-T\cite{shi2023transnext} & 59    & 978   & 51.1  & 51.2  \\
    RMT-S\cite{fan2023rmt} & 56    & 970   & 49.8  & {-} \\
    \rowcolor[rgb]{ 0.8, 1, 1} \textbf{A2Mamba-S} & 60    & 959   & $\mathbf{51.6}$ & $\mathbf{52.0}$ \\
    \midrule
    ConvNeXt-S\cite{liu2022convnet} & 82    & 1027  & 48.7  & 49.6  \\
    SLaK-S\cite{liu2022more} & 91    & 1028  & 49.4  & - \\
    InternImage-S\cite{wang2022internimage} & 80    & 1017  & 50.1  & 50.9  \\
    PeLK-S\cite{chen2024pelk} & 84    & 1077  & 49.7  & - \\
    UniRepLKNet-S\cite{ding2023unireplknet} & 86    & 1036  & 50.5  & 51.0  \\
    MogaNet-B\cite{li2023moganet} & 74    & 1050  & 50.1  & - \\
    % OverLoCK-S & 85    & 1051  & 51.3  & 51.9  \\
    VMamba-S\cite{liu2024vmamba} & 82    & 1038  & 50.6  & 51.2  \\
    MambaVision-S\cite{hatamizadeh2024mambavision} & 84    & 1135  & 48.2  & - \\
    SparX-Mamba-S\cite{lou2025sparx} & 77    & 1039  & 51.3  & 52.5  \\
    Spatial-Mamba-S\cite{xiao2024spatialmamba} & 73    & 992   & 50.6  & 51.4  \\
    Swin-S\cite{liu2021swin} & 81    & 1038  & 47.6  & 49.5  \\
    CSWin-S\cite{dong2022cswin} & 65    & 1027  & 50.4  & 51.5  \\
    UniFormer-B\cite{li2022uniformer} & 80    & 1227  & 50.0  & 50.8  \\
    BiFormer-B\cite{zhu2023biformer} & 88    & 1184  & 51.0  & 51.7  \\
    CAFormer-S36\cite{metaformer2024} & 67    & 1197  & 50.8  & - \\
    TransNeXt-S\cite{shi2023transnext} & 80    & 1089  & 52.2  & 52.3  \\
    RMT-B\cite{fan2023rmt} & 83    & 1111  & 52.0  & - \\
    \rowcolor[rgb]{ 0.8, 1, 1} \textbf{A2Mamba-B} & 80    & 1090  & $\mathbf{53.3}$ & $\mathbf{53.9}$ \\
    \midrule
    SLaK-B\cite{liu2022more} & 135   & 1172  & 50.2  & - \\
    InternImage-B\cite{wang2022internimage} & 128   & 1185  & 50.8  & 51.3  \\
    PeLK-B\cite{chen2024pelk} & 126   & 1237  & 50.4  & - \\
    MogaNet-L\cite{li2023moganet} & 113   & 1176  & 50.9  & - \\
    % OverLoCK-B & 124   & 1202  & 51.7  & 52.3  \\
    VMamba-B\cite{liu2024vmamba} & 122   & 1170  & 51.0	 & 51.6  \\
    MambaVision-B\cite{hatamizadeh2024mambavision} & 126   & 1342  & 49.1  & - \\
    SparX-Mamba-B\cite{lou2025sparx} & 115   & 1181  & 52.3  & 53.4  \\
    Spatial-Mamba-B\cite{xiao2024spatialmamba} & 127   & 1176  & 51.8  & 52.6  \\
    Swin-B\cite{liu2021swin} & 121   & 1188  & 48.1  & 49.7  \\
    CSWin-B\cite{dong2022cswin} & 109   & 1222  & 51.1  & 52.2  \\
    NAT-B\cite{hassani2023neighborhood} & 123   & 1137  & 48.5  & 49.7  \\
    MPViT-B\cite{lee2022mpvit} & 105   & 1186  & 50.3  & - \\
    CAFormer-M36\cite{metaformer2024} & 84    & 1346  & 51.7  & - \\
    TransNeXt-B\cite{shi2023transnext} & 121   & 1268  & 53.0  & 53.4  \\
    RMT-L\cite{fan2023rmt} & 125   & 1324  & 52.8  & - \\
    \rowcolor[rgb]{ 0.8, 1, 1} \textbf{A2Mamba-L} & 126   & 1237  & $\mathbf{53.7}$ & $\mathbf{54.1}$ \\
    \bottomrule
    \end{tabular}%
    }
  \label{tab:upernet}%
\end{table}%

\begin{table*}[t]
  \centering
  \caption{A comparison of semantic segmentation performance among different segmentation models. FLOPs are calculated at 512$\times$512 (ADE20K and COCO-Stuff) and 1024$\times$2048 (Cityscapes) resolutions.}
    \resizebox{0.675\textwidth}{!}{
    \begin{tabular}{l|c|cc|cc|cc}
    \toprule
    \multirow{2}[4]{*}{Method} & \multirow{2}[4]{*}{\# P (M)} & \multicolumn{2}{c|}{ADE20K} & \multicolumn{2}{c|}{Cityscapes} & \multicolumn{2}{c}{COCO-Stuff} \\
\cmidrule{3-8}          &       & \# F (G) & mIoU (\%)  & \# F (G) & mIoU (\%)  & \# F (G) & mIoU (\%) \\
    \midrule
    Segformer-B0\cite{xie2021segformer} & 3.8   & 8.4   & 37.4  & 126   & 76.2  & 8.4   & 35.6  \\
    SegNeXt-T\cite{guo2022segnext} & 4.3   & 7.7   & 41.1  & 62    & 78.9  & 7.7   & 38.7  \\
    VWFormer-B0\cite{yan2024vwformer} & 3.7   & 5.8   & 38.9  & 112   & 77.2  & 5.8   & 36.2  \\
    EDAFormer-T\cite{yu2024embedding} & 4.9   & 5.8   & 42.3  & 152   & 78.7  & 5.8   & 40.3  \\
    CGRSeg-T\cite{ni2024context} & 9.4   & 4.8   & 42.5  & 66    & 78.3  & 4.8   & 40.4  \\
    % CGRSeg-T\cite{ni2024context} & 9.4   & 4.0   & 43.6  & -    & -  & 4.0   & 42.2  \\
    \rowcolor[rgb]{ .867,  .922,  .969} \textbf{SegMAN-T}\cite{fu2025segman} & 6.4   & 6.2   & 43.0  & 53    & 80.3  & 6.2   & 41.3  \\
    \rowcolor[rgb]{ 0.8, 1, 1} \textbf{SegMAN-V2-N} & 6.6   & 7.4   & $\mathbf{44.4}$ & 67    & $\mathbf{81.0}$ & 7.4   & $\mathbf{41.9}$ \\
    \midrule
    % Swin UperNet-T & 60    & 236   & 44.4  & -     & -     & -     & - \\
    ViT-CoMer-S\cite{xia2024vit} & 61    & 296   & 46.5  & -     & -     & -     & - \\
    OCRNet\cite{yuan2020object} & 71    & 165   & 45.6  & -     & -     & -     & - \\
    Segformer-B2\cite{xie2021segformer} & 28    & 62    & 46.5  & 717   & 81.0  & 62    & 44.6  \\
    MaskFormer\cite{cheng2021per} & 42    & 55    & 46.7  & -     & -     & -     & - \\
    Mask2Former\cite{cheng2022masked} & 47    & 74    & 47.7  & -     & -     & -     & - \\
    SegNeXt-B\cite{guo2022segnext} & 28    & 35    & 48.5  & 279   & 82.6  & 35    & 45.8  \\
    FeedFormer-B2\cite{shim2023feedformer} & 29    & 43    & 48.0  & 523   & 81.5  & -     & - \\
    VWFormer-B2\cite{yan2024vwformer} & 27    & 47    & 48.1  & 415   & 81.7  & 47    & 45.2  \\
    EDAFormer-B\cite{yu2024embedding} & 29    & 32    & 49.0  & 606   & 81.6  & 32    & 45.9  \\
    CGRSeg-B\cite{ni2024context} & 36    & 17    & 47.3  & 200   & 80.2  & 17    & 45.2  \\
    LRFormer-S\cite{wu2025lrformer} & 32    & 40    & 50.0  & 295   & 81.9  & 40    & 46.4  \\
    \rowcolor[rgb]{ .867,  .922,  .969} \textbf{SegMAN-S}\cite{fu2025segman} & 29    & 25    & 51.3  & 218   & 83.2  & 25    & 47.5  \\
    \rowcolor[rgb]{ 0.8, 1, 1} \textbf{SegMAN-V2-S} & 32    & 34    & $\mathbf{52.0}$ & 282   & $\mathbf{83.8}$ & 34    & $\mathbf{48.0}$ \\
    \midrule
    Segformer-B3\cite{xie2021segformer} & 47    & 79    & 49.4  & 963   & 81.7  & 79    & 45.5  \\
    SegNeXt-L\cite{guo2022segnext} & 49    & 70    & 51.0  & 578   & 83.2  & 70    & 46.5  \\
    VWFormer-B3\cite{yan2024vwformer} & 47    & 63    & 50.3  & 637   & 82.4  & 63    & 46.8  \\
    LRFormer-B\cite{wu2025lrformer} & 69    & 75    & 51.0  & 555   & 83.0  & 75    & 47.2 \\
    \rowcolor[rgb]{ .867,  .922,  .969} \textbf{SegMAN-B}\cite{fu2025segman} & 52    & 58    & 52.6  & 479   & 83.8  & 58    & 48.4  \\
    \rowcolor[rgb]{ 0.8, 1, 1} \textbf{SegMAN-V2-B} & 56    & 66    & $\mathbf{53.5}$ & 552   & $\mathbf{84.2}$ & 66    & $\mathbf{49.0}$ \\
    \midrule
    % Swin UperNet-B & 121   & 261   & 48.1  & -     & -     & -     & - \\
    ViT-CoMer-B\cite{xia2024vit} & 145   & 455   & 48.8  & -     & -     & -     & - \\
    Segformer-B5\cite{xie2021segformer} & 85    & 110   & 51.0  & 1150  & 82.4  & 110   & 46.7  \\
    VWFormer-B5\cite{yan2024vwformer} & 85    & 96    & 52.0  & 1140  & 82.8  & 96    & 48.0  \\
    LRFormer-L\cite{wu2025lrformer} & 113    & 183    & 52.6  & 908   & 83.2  & 183    & 47.9 \\
    \rowcolor[rgb]{ .867,  .922,  .969} \textbf{SegMAN-L}\cite{fu2025segman} & 92    & 97    & 53.2  & 796   & 84.2  & 97    & 48.8  \\
    \rowcolor[rgb]{ 0.8, 1, 1} \textbf{SegMAN-V2-L} & 108   & 109   & $\mathbf{54.1}$ & 871   & $\mathbf{84.6}$ & 109   & $\mathbf{49.5}$ \\
    \bottomrule
    \end{tabular}%
 }
  \label{tab:segman}%
\end{table*}%

\subsection{Semantic Segmentation}
\textbf{Setup.} We evaluate our backbone architecture (A2Mamba variants) on semantic segmentation using the ADE20K dataset~\cite{zhou2017scene} with the UperNet framework~\cite{xiao2018unified}, following the same training protocol as Swin~\cite{liu2021swin}. Additionally, we assess our segmentation network architecture (SegMAN-V2) on three datasets: ADE20K, Cityscapes~\cite{cordts2016cityscapes}, and COCO-Stuff~\cite{caesar2018cocostuff}, using the same training protocol as SegFormer~\cite{xie2021segformer}. For a fair comparison, all backbone networks are initialized with ImageNet-1K pre-trained weights.
\par
\textbf{Results.} As shown in Table \ref{tab:upernet}, when using the same feature decoder to fairly compare the performance of different backbones, our A2Mamba achieves leading performance compared to other strong baselines. For instance, A2Mamba-S achieves a notable mIoU of 51.6\%, significantly surpassing InternImage-B \cite{wang2022internimage} by 0.8\% and VMamba-B \cite{liu2024vmamba} by 0.6\%, while reducing the number of parameters by about half. This further demonstrates the strong performance of our proposed A2Mamba on dense prediction tasks. On the other hand, when compared with other semantic segmentation models, our previous model, SegMAN \cite{fu2025segman}, has already shown significant performance advantages. However, SegMAN-V2 further improves upon SegMAN, achieving even more significant performance gains. For instance, SegMAN-V2-S has only about one-third of the parameters of Segformer-B5 \cite{xie2021segformer} but achieves 1.0\%, 1.4\%, and 1.3\% higher mIoU on ADE20K, Cityscapes, and COCO-Stuff datasets, respectively. Meanwhile, our SegMAN-V2-B significantly improves LRFormer-B \cite{wu2025lrformer} by 2.5\%, 1.2\%, and 1.8\% on the three datasets, respectively. Furthermore, our SegMAN-V2-L achieves remarkable improvements, outperforming VWFormer-B5 \cite{yan2024vwformer} by 2.1\%, 1.8\%, and 1.5\% on the three datasets, respectively. The consistent performance gains across different datasets and model scales validate the effectiveness of our proposed SegMAN-V2, which can simultaneously capture global contexts, local details, and multi-scale clues through its MASS-based feature encoder and MM-Refine-based feature decoder.

\subsection{Analytical Experiments}
\textbf{Speed comparisons and impact of increased resolution}. Inspired by VMamba~\cite{liu2024vmamba}, we evaluate the inference speed and generalization ability of different vision backbones across various input resolutions. As listed in Table~\ref{tab:speed}, we utilize models pre-trained on ImageNet-1K to perform inference on a range of image resolutions, including $224\times224$, $512\times512$, and $1024\times1024$, and report the corresponding GPU memory consumption (Mem.) and inference throughput (Thr.). The batch sizes used for the three resolutions are 128, 32, and 8, respectively. All experiments are conducted on a single NVIDIA L40S GPU. It can be observed that our proposed A2Mamba achieves competitive efficiency and stronger generalization ability compared to other baselines. For instance, with $224\times224$ inputs, A2Mamba-S outperforms RMT-S in terms of accuracy and achieves $1.5\times$ higher throughput. When the resolution is increased to $512\times512$, A2Mamba-S surpasses RMT-S by a significant margin of $8.5\%$ in top-1 accuracy, while maintaining a speedup of nearly $1.7\times$ and lower memory consumption. Furthermore, when the resolution is extended to $1024\times1024$, A2Mamba-S outperforms RMT-S by a substantial margin of $29.9\%$ in top-1 accuracy, while consuming nearly half the memory and running at $2\times$ speed. Additionally, an interesting phenomenon is that we find advanced vision transformers, such as BiFormer, RMT, and TransNeXt, exhibit significantly increased memory consumption and decreased speed when the resolution is enlarged. This is because, despite the use of efficient attention mechanisms, computational costs still increase significantly at high resolutions. In contrast, our A2Mamba model effectively avoids this phenomenon, owing to its linear-time modules including efficient self-attention and SSM, which enable both efficient computation and memory usage, as well as strong performance, making it a more promising foundation model for complex and high-resolution visual recognition tasks.

\begin{table*}[t]
  \centering
  \caption{Comparison of inference speed and generalization ability over an increasing input resolution.}
  \resizebox{0.975\textwidth}{!}{
    \begin{tabular}{l|c|cccc|cccc|cccc}
    \toprule
    \multirow{2}[4]{*}{Method} & \multirow{2}[4]{*}{\# P (M)} & \multicolumn{4}{c|}{224$\times$224}  & \multicolumn{4}{c|}{512$\times$512}  & \multicolumn{4}{c}{1024$\times$1024} \\
\cmidrule{3-14}          &       & \# F (G) & Mem. (MB) & Thr. (imgs/s) & Acc. (\%) & \# F (G) & Mem.  & Thr. (imgs/s) & Acc. (\%) & \# F (G) & Mem.  & Thr. (imgs/s) & Acc. (\%) \\
    \midrule
    ConvNeXt-T\cite{liu2022convnet}  & 29    & 4.5   & 3263  & 1507  & 82.1  & 23.3  & 3865  & 286   & 78.0  & 93    & 3747  & 70    & 55.4  \\
    ConvNeXt-S\cite{liu2022convnet} & 50    & 8.7   & 3343  & 926   & 83.1  & 45.5  & 3965  & 176   & 80.4  & 182   & 3847  & 43    & 65.4  \\
    ConvNeXt-B\cite{liu2022convnet} & 89    & 15.4  & 4119  & 608   & 83.8  & 80.3  & 4921  & 117   & 80.6  & 321   & 4715  & 28    & 52.9  \\
    \midrule
    FocalNet-T\cite{yang2022focalnet} & 29    & 4.5   & 7151  & 1102  & 82.1  & 23.5  & 9847  & 212   & 78.5  & 94    & 11065  & 52    & 62.2  \\
    FocalNet-S\cite{yang2022focalnet} & 50    & 8.7   & 8679  & 691   & 83.5  & 45.7  & 12685 & 133   & 81.3  & 183   & 15267  & 33    & 67.7  \\
    FocalNet-B\cite{yang2022focalnet} & 89    & 15.4  & 12155  & 477   & 83.8  & 80.6  & 15737 & 88    & 82.9  & 322   & 20858  & 22    & 72.3  \\
    \midrule
    MogaNet-S\cite{li2023moganet} & 25    & 5.0   & 4803  & 766   & 83.8  & 25.9  & 5873  & 145   & 78.7  & 104   & 5831  & 36    & 57.0  \\
    MogaNet-B\cite{li2023moganet} & 44    & 9.9   & 4961  & 373   & 84.3  & 51.7  & 5921  & 70    & 78.2  & 207   & 5967  & 17    & 19.0  \\
    MogaNet-L\cite{li2023moganet} & 83    & 15.9  & 5159  & 282   & 84.7  & 82.9  & 6123  & 53    & 80.2  & 332   & 6053  & 13    & 44.8  \\
    \midrule
    VMamba-T\cite{liu2024vmamba} & 29    & 4.9   & 4663  & 1179  & 82.6  & 25.6  & 5691  & 226   & 80.9  & 103   & 5699  & 56    & 57.4  \\
    VMamba-S\cite{liu2024vmamba} & 50    & 8.7   & 7281  & 596   & 83.6  & 45.5  & 8483  & 115   & 82.9  & 182   & 8915  & 28    & 73.7  \\
    VMamba-B\cite{liu2024vmamba} & 89    & 15.4  & 8767  & 439   & 83.9  & 80.2  & 11035 & 84    & 83.3  & 321   & 11527  & 21    & 74.8  \\
    \midrule
    Swin-T\cite{liu2021swin} & 28    & 4.5   & 4893  & 1324  & 81.3  & 26.6  & 5777  & 213   & 79.0  & 153   & 5521  & 54    & 61.9  \\
    Swin-S\cite{liu2021swin} & 50    & 8.7   & 4961  & 812   & 83.0  & 49.4  & 5865  & 131   & 80.9  & 194   & 5609  & 33    & 65.7  \\
    Swin-B\cite{liu2021swin} & 88    & 15.4  & 6287  & 544   & 83.5  & 87.0  & 7489  & 89    & 81.3  & 342   & 7215  & 22    & 67.7  \\
    \midrule
    MPViT-XS\cite{lee2022mpvit} & 11    & 2.9   & 3511  & 1118  & 80.9  & 15.6  & 4243  & 212   & 78.0  & 62    & 4237  & 48    & 57.2  \\
    MPViT-S\cite{lee2022mpvit} & 23    & 4.7   & 3599  & 808   & 83.0  & 25.1  & 4241  & 153   & 81.1  & 101   & 4269  & 35    & 66.3  \\
    MPViT-B\cite{lee2022mpvit} & 75    & 16.4  & 5981  & 380   & 84.3  & 86.0  & 7431  & 72    & 82.6  & 344   & 7493  & 17    & 66.4  \\
    \midrule
    NAT-M\cite{hassani2023neighborhood} & 20    & 2.7   & 2747  & 1740  & 81.8  & 14.2  & 3191  & 330   & 70.7  & 57    & 3191  & 81    & 38.1  \\
    NAT-T\cite{hassani2023neighborhood} & 28    & 4.3   & 2771  & 1287  & 83.2  & 22.6  & 3227  & 242   & 72.8  & 90    & 3227  & 60    & 39.3  \\
    NAT-S\cite{hassani2023neighborhood} & 51    & 7.8   & 3265  & 823   & 83.7  & 40.8  & 3841  & 156   & 77.1  & 163   & 3841  & 39    & 47.0  \\
    NAT-B\cite{hassani2023neighborhood} & 90    & 13.7  & 4087  & 574   & 84.3  & 71.7  & 4775  & 109   & 78.8  & 287   & 4773  & 27    & 51.6  \\
    \midrule
    BiFormer-T\cite{zhu2023biformer} & 13    & 2.2   & 4567  & 1103  & 81.4  & 16.3  & 7591  & 135   & 71.3  & 117   & 14507  & 21    & 30.0  \\
    BiFormer-S\cite{zhu2023biformer} & 26    & 4.5   & 4635  & 527   & 83.8  & 33.3  & 7645  & 64    & 75.4  & 242   & 14561  & 10    & 40.4  \\
    BiFormer-B\cite{zhu2023biformer} & 57    & 9.8   & 6419  & 341   & 84.3  & 66.9  & 11085 & 42    & 78.0  & 430   & 21761  & 7     & 45.9  \\
    \midrule
    MLLA-T\cite{han2025mlla} & 25    & 4.2   & 4429  & 944   & 83.5  & 21.7  & 5485  & 158   & 81.8  & 87    & 5393  & 37    & 64.8  \\
    MLLA-S\cite{han2025mlla} & 43    & 7.3   & 4505  & 580   & 84.4  & 38.1  & 5561  & 97    & 83.0  & 152   & 5437  & 22    & 69.8  \\
    MLLA-B\cite{han2025mlla} & 96    & 16.2  & 6427  & 341   & 85.3  & 84.5  & 7897  & 57    & 84.0  & 338   & 7885  & 14    & 72.4  \\
    \midrule
    TransNeXt-M\cite{shi2023transnext} & 13    & 2.7   & 4345  & 1054  & 82.5  & 16.3  & 9529  & 94    & 80.9  & 99    & 19793  & 13    & 52.3  \\
    TransNeXt-T\cite{shi2023transnext}& 28    & 5.7   & 5977  & 644   & 84.0  & 33.4  & 13717 & 60    & 82.7  & 185   & 28659  & 9     & 69.6  \\
    TransNeXt-S\cite{shi2023transnext} & 50    & 10.3  & 6069  & 322   & 84.7  & 60.8  & 13779 & 30    & 83.3  & 342   & 29879  & 4     & 71.7  \\
    TransNeXt-B\cite{shi2023transnext} & 90    & 18.4  & 7691  & 225   & 84.8  & 105.1  & 18043 & 22    & 83.8  & 555   & 38633  & 3     & 74.9  \\
    \midrule
    RMT-T\cite{fan2023rmt} & 14    & 2.7   & 3795  & 869   & 82.4  & 18.2  & 6881  & 106   & 74.4  & 131   & 17217  & 13    & 34.2  \\
    RMT-S\cite{fan2023rmt} & 27    & 4.8   & 4689  & 512   & 84.1  & 26.9  & 7035  & 81    & 74.6  & 122   & 10981  & 16    & 42.2  \\
    RMT-B\cite{fan2023rmt} & 54    & 10.4  & 5641  & 260   & 85.0  & 57.7  & 8781  & 42    & 78.5  & 258   & 13675  & 8     & 50.9  \\
    RMT-L\cite{fan2023rmt} & 96    & 19.6  & 7465  & 176   & 85.5  & 106.7  & 11853 & 29    & 80.7  & 463   & 18957  & 6     & 56.6  \\
    \midrule
    \rowcolor[rgb]{ .867,  .922,  .969} \textbf{SegMAN-T Encoder}\cite{fu2025segman} & 4     & 0.7   & 2813  & 2118  & 76.2  & 3.4   & 3325  & 387   & 70.3  & 14    & 3499  & 91    & \textbf{45.7} \\
    \rowcolor[rgb]{ .867,  .922,  .969} \textbf{SegMAN-S Encoder}\cite{fu2025segman} & 26    & 4.1   & 4417  & 708   & 84.0  & 21.4  & 5375  & 139   & 82.4  & 89    & 5401  & 30    & 71.5  \\
    \rowcolor[rgb]{ .867,  .922,  .969} \textbf{SegMAN-B Encoder}\cite{fu2025segman} & 45    & 9.9   & 6551  & 269   & 85.1  & 52.3  & 8247  & 51    & 82.8  & 213   & 8165  & 12    & 72.7  \\
    \rowcolor[rgb]{ .867,  .922,  .969} \textbf{SegMAN-L Encoder}\cite{fu2025segman} & 81    & 16.8  & 6747  & 200   & 85.5  & 88.3  & 8329  & 37    & 81.9  & 357   & 8389  & 9     & 68.1  \\
    \midrule
    \rowcolor[rgb]{ 0.8, 1, 1} \textbf{A2Mamba-N} & 4     & 0.8   & 3273  & 2486  & \textbf{78.7} & 4.4   & 4141  & 445   & \textbf{74.4} & 18    & 3889  & 108   & 43.9  \\
    \rowcolor[rgb]{ 0.8, 1, 1} \textbf{A2Mamba-T} & 15    & 2.7   & 4025  & 1287  & \textbf{83.0} & 14.3  & 5005  & 220   & \textbf{81.4} & 58    & 5921  & 55    & \textbf{66.5} \\
    \rowcolor[rgb]{ 0.8, 1, 1} \textbf{A2Mamba-S} & 31    & 5.4   & 4915  & 762   & \textbf{84.7} & 28.4  & 5935  & 140   & \textbf{83.1} & 117   & 6009  & 32    & \textbf{72.1} \\
    \rowcolor[rgb]{ 0.8, 1, 1} \textbf{A2Mamba-B} & 51    & 10.7  & 6885  & 320   & \textbf{85.7} & 60.2  & 8637  & 60    & \textbf{84.0} & 246   & 8611  & 14    & \textbf{74.8} \\
    \rowcolor[rgb]{ 0.8, 1, 1} \textbf{A2Mamba-L} & 95    & 17.4  & 7905  & 258   & \textbf{86.1} & 91.5  & 9665  & 48    & \textbf{84.6} & 372   & 11825  & 11    & \textbf{75.4} \\
    \bottomrule
    \end{tabular}%
  }
  \label{tab:speed}%
\end{table*}%

\begin{figure}[t]
    \centering
    \includegraphics[width=0.475\textwidth]{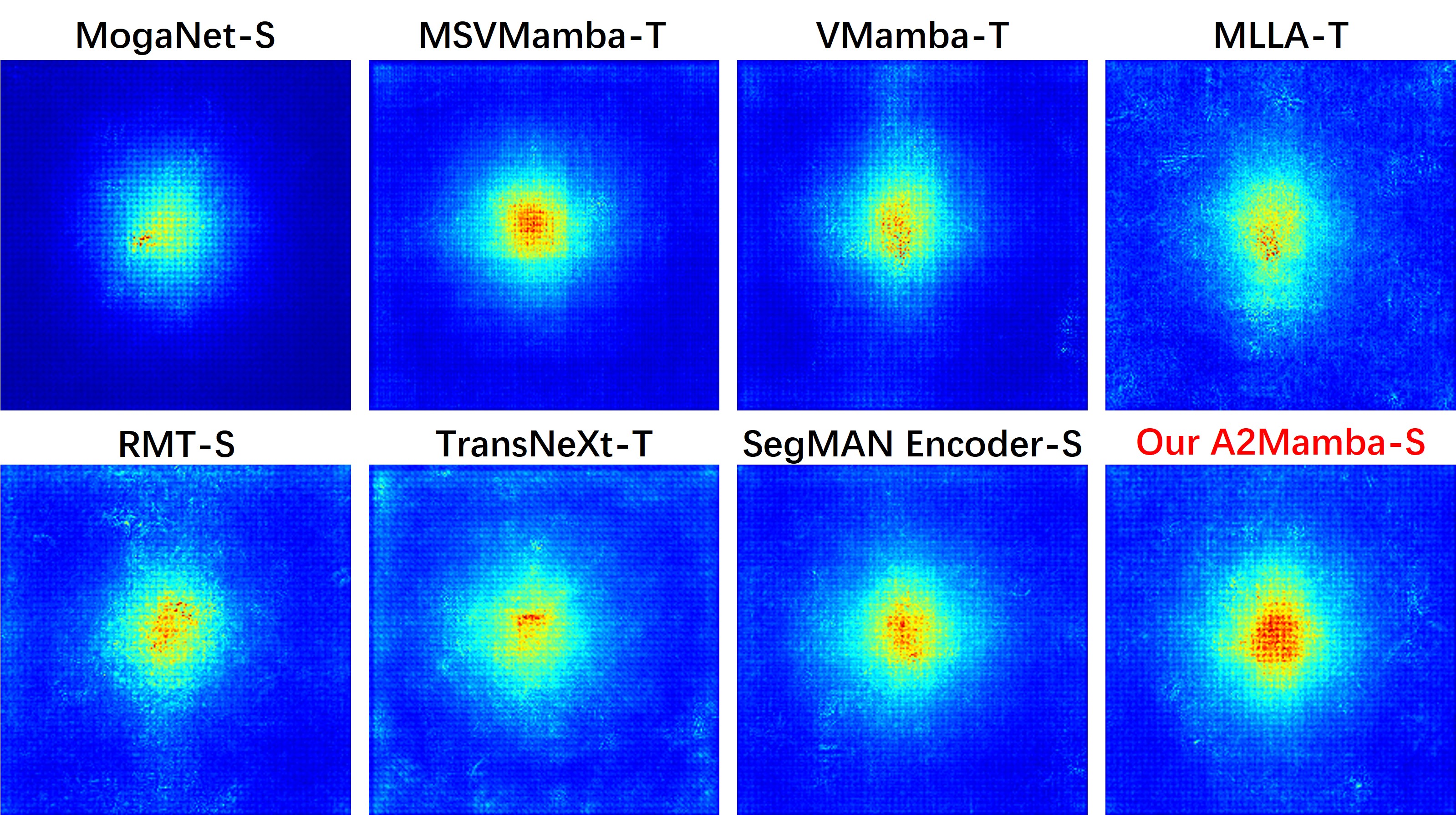}
    \caption{Comparison of ERF among various models.}
    \label{fig:erf}
\end{figure}
\textbf{Effective Receptive Field Analysis}.
To gain further insights into the superiority of A2Mamba over previous methods, we visualize Effective Receptive Fields (ERFs)~\cite{luo2016understanding}. Specifically, we generate the visualizations using over 500 randomly sampled images with a resolution of 224$\times$224 from the ImageNet-1K validation set, while ensuring that all compared models have comparable complexity. As shown in Fig.~\ref{fig:erf}, in comparison to SegMAN Encoder-S using SS2D with four parallel directional scans, our A2Mamba yields a larger ERF, indicating that the attention-augmented SSM can possess stronger global representation capabilities even with a single scan. Furthermore, compared to strong Transformer-based models, including RMT and TransNeXt, our A2Mamba not only exhibits a larger ERF but also demonstrates stronger local sensitivity, benefiting from multi-scale sliding attention. Overall, our A2Mamba model achieves the largest ERF among all strong competitors, including all prior ConvNet-, Transformer-, and Mamba-based models.

\subsection{Ablation Studies}
% TODO: adding Tables for ablation
\textbf{Setup}. We conduct comprehensive ablation studies on image classification and semantic segmentation tasks to evaluate the effectiveness of individual components in our model. Specifically, we train each model variant on the ImageNet-1K dataset for 300 epochs, following the training settings outlined in Section \ref{sec:cls}. Subsequently, we fine-tune the pre-trained models on the ADE20K dataset for 160K iterations with all other settings identical to those of SegFormer \cite{xie2021segformer}. Unless otherwise stated, the segmentation networks are built upon our MM-Refine-based decoder. FLOPs and throughput are measured at 512$\times$512 image resolution with a batch size of 32 using the backbone on a single NVIDIA L40S GPU, following the protocol of \cite{fu2025segman}.
% \#T refers to the inference throughput of a model on an NVIDIA L40S GPU with a batch size of 128 and an image size of 224$\times$224.
\par
\textbf{A roadmap from LASS to MASS}. We provide a detailed evolution of the LASS mixer~\cite{fu2025segman} towards the MASS mixer. As listed in Table~\ref{tab:roadmap}, we first replace all MASS mixers in the A2Mamba-T model with the LASS mixer, resulting in our baseline model with 82.2\% top-1 accuracy and 48.2\% mIoU, respectively. Then, we substitute Natten \cite{hassani2023neighborhood} in LASS with our Adaptive Multi-scale Attention (AMA) discussed in Section~\ref{sec:mixer}, yielding 0.3\%/0.5\% improvement in top-1/mIoU. This highlights the importance of adaptive multi-scale modeling, particularly in semantic segmentation tasks. Next, we replace SS2D~\cite{liu2024vmamba} with vanilla SSM~\cite{gu2023mamba}, which leads to a significant performance drop with 81.4\% top-1 accuracy and 47.3\% mIoU. This suggests that using only unidirectional scanning severely impairs the model's ability to capture the contextual information of an input image. However, when we replace SSM with the proposed A2SSM discussed in Section~\ref{sec:mixer}, the performance improves substantially by 1.3\%/1.9\% in top-1/mIoU, demonstrating the strong spatial perception and dynamic capabilities of our A2SSM. Finally, we introduce a gating mechanism~\cite{gu2023mamba,li2023moganet} to the model, which results in the final version of our MASS mixer, achieving both improved performance and efficiency compared to the baseline model.

\begin{table}[t]
  \centering
  \caption{
  A detailed roadmap that incrementally evolves LASS \cite{fu2025segman} to our proposed MASS. 
  }
    \resizebox{0.475\textwidth}{!}{
    \begin{tabular}{lccccc}
    \toprule
    Model & \# P (M) & \# F (G) & Thr. (imgs/s) & Acc. (\%) & mIoU (\%) \\
    \midrule
    Baseline & 13    & 14.5  & 176   & 82.2  & 48.2  \\
    Natten $\rightarrow$ AMA & 13    & 14.5  & 172   & 82.5  & 48.7  \\
    SS2D $\rightarrow$ SSM & 13    & 12.0  & 256   & 81.4  & 47.3  \\
    SSM $\rightarrow$ A2SSM & 13    & 13.3  & 235   & 82.7  & 49.2  \\
    \rowcolor[rgb]{ .867,  .922,  .969}w Gate & 15    & 14.0  & 220   & \textbf{83.0} & \textbf{49.7} \\
    \bottomrule
    \end{tabular}%
   }
  \label{tab:roadmap}%
\end{table}%
\par
\textbf{Impact of dual-branch design and dilation rates}. We investigate the impact of the dual-branch design and the dilation rate ($\mathbf{r}$) of AMA on model performance. The baseline model is A2Mamba-T, which uses a dual-branch AMA with an adaptive dilation rate as described in Equation~\ref{eq:dil}. First, we set the dilation rates to fixed values, namely 3, 5, and 7, respectively, while keeping the dual-branch structure unchanged. 
As shown in Table~\ref{tab:dil}, it is evident that using a fixed $\mathbf{r}$ has a negligible impact on image classification performance, but leads to a significant decline in semantic segmentation performance. 
Second, to examine whether the dual-branch structure itself is essential to MASS, we further degrade AMA into two single-branch variants of A2Mamba-T, in which a single sliding attention with either dilation=1 or an adaptive dilation rate interacts directly with the SSM. 
As shown in Table~\ref{tab:dil}, although removing the dual-branch design degrades performance, both single-branch variants still achieve competitive top-1 accuracy (82.6\% and 82.4\%) and mIoU (48.8\% and 48.6\%), with higher inference speed than the dual-branch baseline. 
Between the two, the variant retaining only the local branch (dilation=1) slightly outperforms the one retaining only the dilated branch, because the former preserves fine-grained local details while the SSM continues to supply global context. In contrast, the adaptive dilation rate enlarges the attention scope towards a sparse global field and therefore loses some capacity to extract fine-grained local details. 
These results indicate that the two branches of AMA play complementary roles, confirming that the dual-branch design is a beneficial architectural option that further strengthens multi-scale modeling, rather than the essential ingredient of MASS. 
Finally, we modify the dual-branch AMA to a four-branch version, where one branch is regular sliding local attention, and the remaining three branches are dilated local attention with $\mathbf{r}=\left \{ 3, 5, 7 \right \}$, respectively. However, this modification does not bring about performance improvements and instead reduces efficiency. These results demonstrate that adaptively adjusting the dilation rate according to the input resolution can capture more useful multi-scale information in dense predictions.
\par
\begin{table}[t]
  \centering
  \caption{An investigation of the dual-branch design and dilation rates in AMA.}
  \resizebox{0.475\textwidth}{!}{
    \begin{tabular}{lcccll}
    \toprule
    Model & \# P (M) & \# F (G) & Thr. (imgs/s) & Acc. (\%) & mIoU (\%) \\
    \midrule
    \multicolumn{6}{c}{Dual-branch} \\
    \rowcolor[rgb]{ .867,  .922,  .969}Baseline & 15    & 14    & 220   & \textbf{83.0} & \textbf{49.7} \\
    Dilation=3 & 15    & 14    & 221   & 82.8$(-0.2)$ & 49.1$(-0.6)$ \\
    Dilation=5 & 15    & 14    & 221   & 83.0$(+0.0)$ & 49.3$(-0.4)$ \\
    Dilation=7 & 15    & 14    & 221   & 83.0$(+0.0)$ & 49.2$(-0.5)$ \\
    \multicolumn{6}{c}{Single-branch} \\
    Local branch only (Dilation=1) & 15    & 14    & 229   & 82.6$(-0.4)$ & 48.8$(-0.9)$    \\
    Dilated branch only (Adaptive Dilation) & 15    & 14    & 225   & 82.4$(-0.6)$ & 48.6$(-1.1)$    \\
    \multicolumn{6}{c}{Multi-branch} \\
    Dilation=$\left \{ 3, 5, 7 \right \}$ & 15    & 14    & 209   & 82.8$(-0.2)$ & 49.5$(-0.2)$ \\
    \bottomrule
    \end{tabular}%
   }
  \label{tab:dil}%
\end{table}%

\textbf{Impact of shared attention maps}. The core of our A2SSM is using a variant of cross-attention with shared multi-scale attention maps to efficiently enhance the spatial perception and dynamic modeling capabilities of SSM. To verify this, we take A2Mamba-T as the baseline model and replace the cross-attention operation with other related operations, including dilated RepConv~\cite{ding2023unireplknet} and DCNv2~\cite{zhu2019deformable}. To ensure a fair comparison, we use the same kernel size as the original local attention window size for dilated RepConv and DCNv2. Note that we use the depthwise version of DCNv2, as the original version incurs significant computational costs. As listed in Table~\ref{tab:ssm_local}, using either dilated RepConv or DCNv2 results in significant performance and efficiency degradation. This is because these operators cannot dynamically capture the multi-scale relationships among tokens, leading to ineffective spatial structure perception and dynamic enhancement when embedded into SSM.
Furthermore, to examine whether the attention maps contribute through their spatially meaningful correspondence rather than through the mere addition of an operation, we keep the entire A2Mamba-T unchanged but randomly permute the spatial positions of the multi-scale attention maps before they are used to aggregate the hidden states of the SSM. As shown in Table~\ref{tab:ssm_local}, this shuffled-attention variant drops performance on both tasks significantly, especially on semantic segmentation ($-2.2\%$ mIoU). This is because dense prediction requires position-accurate representations at every token and is therefore highly sensitive to spatial misalignment, whereas image classification aggregates information into a single category prediction and is comparatively more tolerant of such perturbations. Once the attention maps are shuffled, the cross-attention step aggregates each SSM hidden state over an incorrect multi-scale neighborhood, so the position-accurate spatial structure that A2SSM is designed to strengthen is instead corrupted. These results confirm that the effectiveness of A2SSM originates from the position-aware attention maps rather than from simply attaching an extra operation to the SSM.

\begin{table}[t]
  \centering
  \caption{Effect of different spatial aggregators and shuffled attention maps on SSM.}
  \resizebox{0.475\textwidth}{!}{
    \begin{tabular}{lccccc}
    \toprule
    Model & \# P (M) & \# F (G) & Thr. (imgs/s) & Acc. (\%) & mIoU (\%) \\
    \midrule
    \rowcolor[rgb]{ .867,  .922,  .969}Baseline & 15    & 14.0  & 220   & \textbf{83.0} & \textbf{49.7} \\
    Dilated RepConv\cite{ding2023unireplknet} & 16    & 13.9  & 201   & 81.9  & 47.9  \\
    DCNv2\cite{zhu2019deformable} & 16    & 14.6  & 93    & 82.1  & 48.3  \\
    Shuffled Attn & 15    & 14.0  & 213   & 81.5  & 47.5 \\
    \bottomrule
    \end{tabular}%
   }
  \label{tab:ssm_local}%
\end{table}%

\par
\textbf{A comparison of token mixers}. Following our conference version~\cite{fu2025segman}, we replace the token mixer in the SegMAN-S encoder with those of other vision backbones, including PVT~\cite{wang2021pyramid}, MaxViT~\cite{tu2022maxvit}, ACmix~\cite{pan2022integration}, and BiFormer~\cite{zhu2023biformer}, to conduct a fair comparison of different token mixers. As shown in Table~\ref{tab:mixer_compare}, our MASS token mixer achieves notable performance improvements on both classification and segmentation tasks, while maintaining competitive computational costs. The performance gains can be attributed to the complementary nature of our approach, which models adaptive multi-scale clues and more robust global contexts.
\par

\begin{table}[t]
  \centering
  \caption{A comparison of different token mixers.}
  \resizebox{0.475\textwidth}{!}{
    \begin{tabular}{lccccc}
    \toprule
    Token Mixer & \# P (M) & \# F (G) & Thr. (imgs/s) & Acc. (\%) & mIoU (\%) \\
    \midrule
    PVT\cite{wang2021pyramid}   & 30    & 22.0  & 169   & 82.8  & 49.1  \\
    MaxViT\cite{tu2022maxvit} & 25    & 29.8  & 96    & 83.5  & 47.2  \\
    ACmix\cite{pan2022integration} & 25    & 19.3  & 104   & 83.1  & 48.6  \\
    BiFormer\cite{zhu2023biformer}   & 25    & 30.5  & 97    & 82.9  & 48.8  \\
    LASS\cite{fu2025segman}  & 26    & 21.4  & 139   & 84.0  & 51.3  \\
    \rowcolor[rgb]{ .867,  .922,  .969}MASS  & 27    & 22.8  & 160   & \textbf{84.3} & \textbf{51.8} \\
    \bottomrule
    \end{tabular}%
   }
  \label{tab:mixer_compare}%
\end{table}%

\begin{table}[!t]
  \centering
  \caption{A detailed roadmap that incrementally evolves the SegMAN decoder to our SegMAN-V2 decoder.}
    \resizebox{0.475\textwidth}{!}{
    \begin{tabular}{lcccc}
    \toprule
    Model & \# P (M) & \# F (G) & Thr. (imgs/s) & mIoU (\%) \\
    \midrule
    MMSCopE\cite{fu2025segman} & 17    & 18.1  & 142   & 48.5  \\
    w Progressive Down. & 16    & 17.1  & 150   & 48.8  \\
    w Local Embed. (\textit{k}=3) & 17    & 17.2  & 147   & 48.9  \\
    w Local Embed. (\textit{k}=5) & 17    & 17.2  & 143   & 49.1  \\
    w Local Embed. (\textit{k}=7) & 17    & 17.2  & 136   & 49.1  \\
    w MASS & 18    & 17.1  & 140   & 49.5  \\
    \rowcolor[rgb]{ .867,  .922,  .969}{w Low Level} & 18    & 17.6  & 137   & \textbf{49.7} \\
    \bottomrule
    \end{tabular}%
   }
  \label{tab:seg_road}%
\end{table}%

\textbf{A roadmap from SegMAN decoder to SegMAN-V2 decoder}. SegMAN-V2 decoder is an upgraded version of SegMAN decoder~\cite{fu2025segman}, aiming to achieve more fine-grained semantic segmentation. To this end, we provide a detailed roadmap to illustrate the performance improvements of our SegMAN-V2 decoder. All experiments are conducted using A2Mamba-T as the encoder on the ADE20K dataset, following the same training settings as SegFormer~\cite{xie2021segformer}. FLOPs and throughput of a segmentation network are evaluated using 512$\times$512 input resolution with a batch size of 32 on a single NVIDIA L40S GPU, following the setup of \cite{fu2025segman}. As listed in Table~\ref{tab:seg_road}, we first modify the original downsampling in MMSCopE to a more progressive downsampling described in Section~\ref{sec:seg}, resulting in improved performance and efficiency. Next, we introduce a local embedding based on dilated RepConv to supplement the lost local details, and our experiments show that a kernel size of 5$\times$5 (\textit{k}=5) achieves the optimal trade-off. Subsequently, we replace SS2D with the MASS mixer, leading to further significant performance improvements. Finally, we employ low-level enhancement, which yields modest performance gains without obviously compromising efficiency.
\par
\textbf{A full factorial ablation of the encoder and decoder}. The two roadmaps above already provide two controlled comparisons that respectively isolate the encoder and the decoder: Table~\ref{tab:roadmap} keeps the decoder at MM-Refine and varies the encoder from LASS to MASS, while Table~\ref{tab:seg_road} presents A2Mamba (MASS) as the encoder and varies the decoder from MMSCopE to MM-Refine. To further complete this isolation, we additionally train the one remaining combination, in which the encoder architecture is kept identical to A2Mamba-T but every MASS mixer is replaced by the LASS mixer~\cite{fu2025segman}, paired with the original MMSCopE decoder. Together with the results already reported above, Table~\ref{tab:factorial} summarizes all four encoder-decoder combinations. When the decoder is fixed, replacing LASS with MASS improves mIoU by 1.1\% under MMSCopE and by 1.5\% under MM-Refine. When the encoder is fixed, replacing MMSCopE with MM-Refine improves mIoU by 0.8\% under LASS and by 1.2\% under MASS. Notably, combining MASS with MM-Refine improves mIoU by 2.3\% over the LASS-MMSCopE combination used in our conference version. This demonstrates that the encoder-decoder segmentation network formed by our A2Mamba and MM-Refine has a strong capability in modeling multi-scale and global contexts, achieving a significant improvement. Moreover, such performance gains are obtained with better efficiency, as A2Mamba with MM-Refine achieves a higher inference speed than the SegMAN encoder with MMSCopE.
\par

\begin{table}[t]
  \centering
  \caption{A full factorial comparison isolating the contributions of the encoder (mixer) and the decoder. Note that the encoders use the same network layout for fair comparisons.}
  \resizebox{0.475\textwidth}{!}{
    \begin{tabular}{llcccc}
    \toprule
    Encoder (Mixer) & Decoder & \# P (M) & \# F (G) & Thr. (imgs/s) & mIoU (\%) \\
    \midrule
    SegMAN Encoder (LASS) & MMSCopE & 15 & 18.6 & 122 & 47.4 \\
    SegMAN Encoder (LASS) & MM-Refine & 16 & 18.1 & 119 & 48.2 \\
    A2Mamba (MASS) & MMSCopE & 17 & 18.1 & 142 & 48.5 \\
    \rowcolor[rgb]{ .867,  .922,  .969}A2Mamba (MASS) & MM-Refine & 18 & 17.6 & 137 & \textbf{49.7} \\
    \bottomrule
    \end{tabular}%
   }
  \label{tab:factorial}%
\end{table}%

\section{Conclusion}
This work presents A2Mamba, a robust Transformer-Mamba hybrid vision backbone architecture, which features a unified token mixer dubbed Multi-scale Attention-augmented State Space Model (MASS). The MASS module adaptively extracts multi-scale contexts, while storing interim attention maps for further enhancing the global perception and dynamic modeling capabilities of the subsequent SSM layer. We evaluate A2Mamba on diverse vision tasks, including image classification and dense predictions, and demonstrate its significant performance advantages over existing strong ConvNet-, Transformer-, and Mamba-based vision backbone architectures.

\bibliography{ref}
\bibliographystyle{ieeetr}

\end{document}